%% file: colm2026_conference.tex
\documentclass{article} % For LaTeX2e
\usepackage[preprint]{colm2026_conference}

\usepackage{microtype}
\usepackage{hyperref}
\usepackage{url}
\usepackage{booktabs}

\usepackage{algorithm}
\usepackage{algorithmic}
\usepackage{amsmath}
\usepackage{amsthm}
\usepackage{amssymb}
\usepackage{array}
\usepackage{color}
\usepackage{caption}
\usepackage{mathtools}
\usepackage{mathrsfs}
\usepackage{wrapfig}
\usepackage{multirow}
\usepackage{diagbox}
\usepackage{bm}
\usepackage{tablefootnote}
\usepackage{fancybox}
\usepackage{colortbl}
\usepackage{subcaption}
\usepackage{comment}
\usepackage{fvextra}

\definecolor{SDEblue}{RGB}{28 58 88}
\definecolor{cc1}{rgb}{1.0, 0.44, 0.37}
\definecolor{cc2}{rgb}{0.0, 0.2, 0.6}
\definecolor{cc3}{RGB}{255, 191, 0}
\definecolor{cc4}{RGB}{0, 128, 128}
\hypersetup{
 colorlinks=true,
 urlcolor=magenta,
 citecolor=SDEblue,
}

\usepackage{tcolorbox}
\definecolor{lightroyalblue}{HTML}{F6F8FD} % more blue: E5EAFB
\definecolor{royalblue}{HTML}{4169E1}
\definecolor{lighterblue}{HTML}{f2fafd}  % more blue: e4f4fa
\newtcolorbox{abox}{colback=lightroyalblue,colframe=black}
\definecolor{LightCyan}{rgb}{.9, .95, 1.}

\usepackage{lineno}

\definecolor{darkblue}{rgb}{0, 0, 0.5}
\hypersetup{colorlinks=true, citecolor=darkblue, linkcolor=darkblue, urlcolor=darkblue}

\title{Understanding Performance Gap Between Parallel and \\ Sequential Sampling in Large Reasoning Models}

\renewcommand\footnotemark{}
\author{
Xiangming Gu$^{*1,2}$, Soham De$^{1}$, Larisa Markeeva$^{1}$, 
Petar Veli\v{c}kovi\'{c}$^{1}$, Razvan Pascanu$^{1}$\\[0.5em]
$^{1}$Google DeepMind \quad $^{2}$National University of Singapore\\
\thanks{$^*$Work performed while the author was at Google DeepMind and finished before September 2025.}
\footnotesize{\texttt{\{xiangming, sohamde, lmarkeeva, 
petarv, razp\}@google.com}} \\
}

\begin{document}

\ifcolmsubmission
\linenumbers
\fi

\maketitle

\input{sections/1_abstract}
\input{sections/2_introduction}

\input{sections/3_preliminary}
\input{sections/4_main}

\bibliography{colm2026_conference}
\bibliographystyle{colm2026_conference}

\appendix
\input{sections/TBD}

\end{document}

%% file: sections/1_abstract.tex
\begin{abstract}
    Large Reasoning Models (LRMs) have shown remarkable performance on challenging questions, such as math and coding. However, to obtain a high quality solution, one may need to sample more than once. In principal, there are two sampling strategies that can be composed to form more complex processes: sequential sampling and parallel sampling. In this paper, we first compare these two approaches with rigor, and observe, aligned with previous works, that parallel sampling seems to outperform sequential sampling even though the latter should have more representation power. To understand the underline reasons, we make three hypothesis on the reason behind this behavior: (i) parallel sampling outperforms due to the aggregator operator; (ii) sequential sampling is harmed by needing to use longer contexts; (iii) sequential sampling leads to less exploration due to conditioning on previous answers. The empirical evidence on various model families and sizes (Qwen3, DeepSeek-R1 distilled models, Gemini 2.5) and question domains (math and coding) suggests that the aggregation and context length do not seem to be the main culprit behind the performance gap. In contrast, the lack of exploration seems to play a considerably larger role, and we argue that this is one main cause for the performance gap.\looseness=-1
\end{abstract}

%% file: sections/2_introduction.tex
\vspace{-0.2cm}
\section{Introduction}
\vspace{-0.2cm}

Recently, Large Reasoning Models (LRMs)~\citep{jaech2024openai,guo2025deepseek,comanici2025gemini,anthropic2025claude4,xai2025grok3,yang2025qwen3} significantly advance the frontier of reasoning abilities. By scaling the thinking traces, LRMs demonstrate improved performance. However, such scaling cannot persist as there is still an upper limit for thinking tokens. Additionally, the literature~\citep{gema2025inverse,ghosal2025does} demonstrates that further extending thinking trajectories may result in inverse test-time-scaling or overthinking.

This raises a question: \emph{When LRMs are set to full potential (unlimited thinking budget), what are reliable ways to further scale performance?} Previous literature have already shown that scaling number of sampled solutions can bring significant performance gains. This includes either sampling solutions in parallel and then aggregate them, such as majority voting~\citep{wang2022self} and best-of-N~\citep{snell2024scaling,li2022competition}, or sampling solutions in a sequence chain with a self-refinement manner~\citep{madaan2023self,chen2023teaching,huang2023large,tian2025think}. The above two sampling approaches are termed as \emph{parallel sampling} and \emph{sequential sampling} in this paper. The essential distinction is that given a question to solve, parallel sampling generates solution candidates independently while sequential sampling generates each solution based on previous solutions. Additionally, parallel sampling adopts an aggregation operator to reach the final solution while sequential sampling always takes the solution in the final round (see Appendix~\ref{related} for a detailed discussion on related work).

In this paper, we firstly conduct empirical studies to compare the scaling properties of sequential and parallel sampling on competition math and coding scenarios. Our results indicate that sequential sampling is suboptimal compared to parallel sampling under the same number of solutions even though the former has more representation power in theory. Therefore, in this paper, our focus is to understand why parallel sampling has such superiority compared to sequential sampling.  Based on the differences between sequential and parallel sampling, we propose three hypothesis about the underlying reason: (i) parallel sampling outperforms due to the aggregator operator; (ii) sequential sampling is harmed by needing to use longer contexts; (iii) sequential sampling leads to less exploration due to conditioning on previous answers. To validate these hypothesis, we conduct a series of experiments using different model families and scales, including Qwen3~\citep{yang2025qwen3}, DeepSeek-R1 distilled models~\citep{guo2025deepseek}, Gemini 2.5~\citep{comanici2025gemini}, for both math competition and coding generation tasks. Our empirical evidence suggests that the aggregation and context length are not the main reason. In contrast, we argue that the lack of exploration seems to play a considerably role in the performance gap between parallel and sequential sampling.\looseness=-1

% (i) parallel sampling adopts aggregation approach, which is absent in sequential sampling; (ii) sequential sampling has longer input context, resulting in performance degradation; (iii) with previous solutions in the context, the solution exploration in sequential sampling is reduced. Through empirical evidence, we find the third hypothesis is the most convincing one to explain the performance gap between parallel and sequential sampling in LRMs.\looseness=-1

% Beyond the above main claim in this paper, we also have the following takeaways:
% \begin{itemize}
%     \item With more previous solutions in the context, 
% \end{itemize}

% LRMs are either prompted to sample longer thinking traces or sample multiple trials based on previous solutions through a self-refinement approach. However, the lengths of the thinking traces are difficult to control, especially when the target length exceeds the maximum thinking budgets.  Therefore, we are most interested in the sequential self-refinement trials in this paper. 

%% file: sections/3_preliminary.tex
\vspace{-0.1cm}
\section{Preliminaries on Parallel/Sequential Sampling}
\vspace{-0.1cm}

In this section, we provide notations to describe parallel and sequential sampling for Large Reasoning Models (LRMs).  We denote the policy of LRM as $\pi_{\theta}(\cdot)$ and the prompt as $\boldsymbol{p}$, then the thinking trace can be represented as $\boldsymbol{z}\sim \pi_{\theta}(\cdot|\boldsymbol{p})$,  and the solution is $\boldsymbol{y}\sim \pi_{\theta}(\cdot|\boldsymbol{p}\textrm{,}\,\boldsymbol{z})$.\looseness=-1

\textbf{Parallel sampling.} We independently sample $N$ multiple thinking traces and solutions: $\boldsymbol{z}_n\sim \pi_{\theta}(\cdot|\boldsymbol{p})\textrm{,}\,\boldsymbol{y}_n\sim \pi_{\theta}(\cdot|\boldsymbol{p}\textrm{,}\,\boldsymbol{z}_n)\textrm{,}\,\,\,n=1\textrm{,}\,2\textrm{,}\,\cdots\textrm{,}\,N$. Then an aggregation operator $\varphi(\cdot)$ is adopted to select the final solution: $\boldsymbol{y}^*=\varphi(\boldsymbol{y}_1\textrm{,}\,\boldsymbol{y}_2\textrm{,}\,\cdots\textrm{,}\,\boldsymbol{y}_N)$.

% Ideally, different trials share the same input prompt prefilling, so the total token counts will be $|\boldsymbol{p}|+\sum_{n=1}^N(|\boldsymbol{z}_n|+|\boldsymbol{y}_n|)$. 

\begin{figure}[t]
\centering
\includegraphics[width=\textwidth]{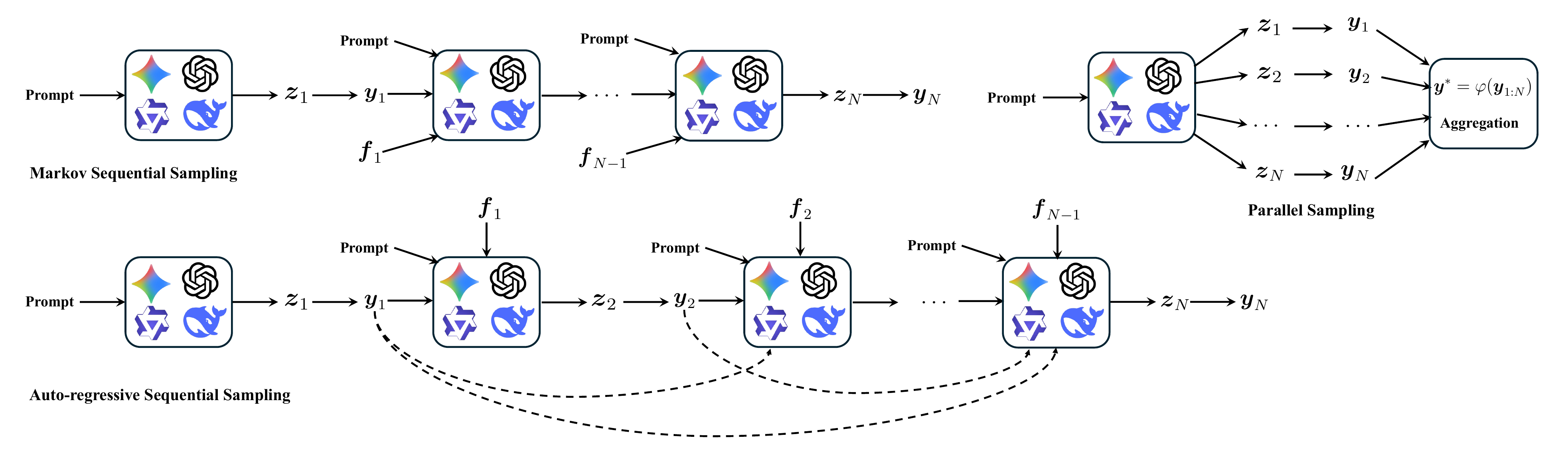}
\vspace{-0.2cm}
\caption{Illustration of sampling approaches considered in this work. In parallel sampling, given a prompt, multiple independent thinking traces and solutions are sampled, and then the solutions are aggregated to the final one. In Markov sequential sampling, solutions are sampled in a sequential chain, for current generation, only last solution is considered as input. While for auto-regressive sequential sampling, the input includes all previous solutions.\looseness=-1}
\vspace{-0.4cm}
\label{illustration}
\end{figure}

\textbf{Sequential sampling.} The $n$-th thinking trace and solution depend on previous sampling output. To formalize such a dependence, we first have the initial sample $\boldsymbol{z}_1\sim \pi_{\theta}(\cdot|\boldsymbol{p})\textrm{,}\,\boldsymbol{y}_1\sim \pi_{\theta}(\cdot|\boldsymbol{p}\textrm{,}\,\boldsymbol{z}_1)$. Then for $n>1$\looseness=-1
\begin{align}
    \boldsymbol{z}_n&\sim \pi_{\theta}(\cdot|\boldsymbol{p}\textrm{,}\,\textrm{Context}(\boldsymbol{z}_1\textrm{,}\,\boldsymbol{y}_1\textrm{,}\,\cdots\textrm{,}\,\boldsymbol{z}_{n-1}\textrm{,}\,\boldsymbol{y}_{n-1}))\textrm{,}\\\nonumber
    \boldsymbol{y}_n&\sim \pi_{\theta}(\cdot|\boldsymbol{p}\textrm{,}\,\textrm{Context}(\boldsymbol{z}_1\textrm{,}\,\boldsymbol{y}_1\textrm{,}\,\cdots\textrm{,}\,\boldsymbol{z}_{n-1}\textrm{,}\,\boldsymbol{y}_{n-1})\textrm{,}\,\boldsymbol{z}_n)\textrm{,}\nonumber
\end{align}
here $\textrm{Context}(\cdot)$ refers to the operation on managing previous samples in the LRM context. Since thinking traces are very long, they can be eliminated from the history in multi-turn interactions, e.g., \citet{jaech2024openai,yang2025qwen3}. Alternatively, Gemini 2.5~\citep{comanici2025gemini} compress thinking traces into thought summaries. Between two samples, there is another feedback prompt $\boldsymbol{f}$  promoting LRMs to enter the next round of sampling. Therefore, $\textrm{Context}(\cdot)$ can be expressed as: $\textrm{Context}(\boldsymbol{z}_1\textrm{,}\,\boldsymbol{y}_1\textrm{,}\,\cdots\textrm{,}\,\boldsymbol{z}_{n-1}\textrm{,}\,\boldsymbol{y}_{n-1})=\{\boldsymbol{z}'_1\textrm{,}\,\boldsymbol{y}'_1\textrm{,}\,\boldsymbol{f}_1\textrm{,}\,\cdots\textrm{,}\,\boldsymbol{z}'_{n-1}\textrm{,}\,\boldsymbol{y}'_{n-1}\textrm{,}\,\boldsymbol{f}_{n-1}\}\textrm{,}$ where $\boldsymbol{z}'_{n}$ can be nothing or a compressed version of $\boldsymbol{z}_{n}$, and $\boldsymbol{y}'_{n}$ is equal to $\boldsymbol{y}_{n}$ or the extracted answer, e.g., coding program. We term the above context management \textbf{auto-regressive} sequential sampling, which arranges full history in the context. Inspired by \citet{tian2025think}, we also consider a simpler context management called (first-order) \textbf{Markov} sequential sampling, which only considers the last history: $\textrm{Context}(\boldsymbol{z}_1\textrm{,}\,\boldsymbol{y}_1\textrm{,}\,\cdots\textrm{,}\,\boldsymbol{z}_{n-1}\textrm{,}\,\boldsymbol{y}_{n-1})=\{\boldsymbol{z}'_{n-1}\textrm{,}\,\boldsymbol{y}'_{n-1}\textrm{,}\,\boldsymbol{f}_{n-1}\}$ by re-starting the chat: ``Here is your previous response:\textbackslash{}n\{LAST\_SOLUTION\}\textbackslash{}n''. For coding question, we could only include the coding program: ``Here is the code in your previous response:\textbackslash{}n\{LAST\_CODE\}\textbackslash{}n''.\looseness=-1

\vspace{-0.2cm}
\section{Empirical Comparison between Parallel/Sequential Sampling}
\vspace{-0.2cm}

% In this section, we consider challenging math and coding competitions as testbed. The explored LRMs include Qwen3~\citep{yang2025qwen3}, DeepSeek-R1-Distill Qwen~\citep{guo2025deepseek}, and Gemini 2.5 families~\citep{comanici2025gemini}. 

% \begin{figure}[t]
% \centering
% \includegraphics[width=\textwidth]{figures/aime2.pdf}
% \caption{Compare performances of (\emph{Left}) Gemini 2.5-flash (\emph{Middle}) Qwen3-8B (\emph{Right}) Qwen3-14B using parallel sampling and sequential sampling on AIME 2025. We use pass@k aggregation.\looseness=-1}
% % \vspace{-0.2cm}
% \label{math_compare2}
% \end{figure}

\subsection{Math Competition}

\textbf{Setups.} We mainly use the AIME2025 benchmark~\citep{aops2025aime}, which includes 30 questions. The ground-truth answer for each question is an integer between 0 and 999. The model answer (extracted from answer box) is considered to be correct if it matches the ground-truth answer. For parallel sampling, we sample 64 solutions and then randomly sample certain number of solutions for aggregation. We adopt majority voting~\citep{wang2022self} as the aggregation operator. For sequential sampling, we sample 8 rounds in sequence and then conduct 8 repeated experiments to compute the accuracy. We consider two feedback prompts: ``Please re-answer.'' and ``Please review your previous response and find problems with your answer. Based on the problems you found, improve your answer.'' (self-refinement feedback) to promote LRMs reflect the previous solutions and generate a new one, inspired by \citet{tian2025think,huang2023large}. We use the generation hyper-parameters according to each model's best recommendation.

\begin{figure}[t]
\centering
\vspace{-0.2cm}
\includegraphics[width=\textwidth]{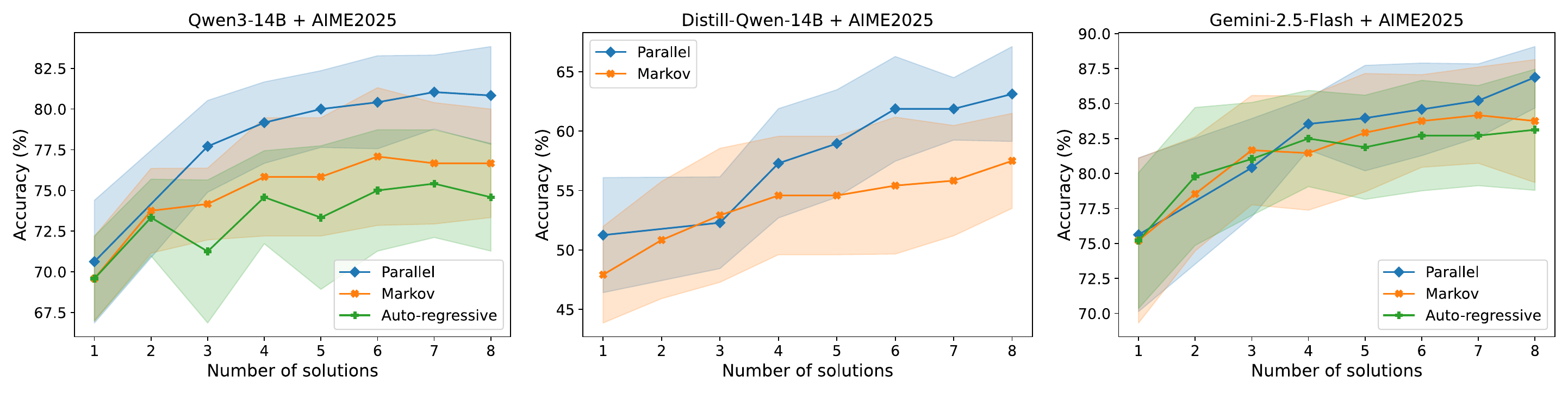}
\vspace{-0.2cm}
\caption{Comparisons of AIME2025 performance between parallel and sequential sampling using (\emph{Left}) Qwen3-14B; (\emph{Middle}) DeepSeek-R1-Distill Qwen-14B; (\emph{Right}) Gemini 2.5 Flash. In parallel sampling, majority voting aggregation is adopted. Parallel sampling performs better than sequential sampling on the math competition scenario.\looseness=-1}
\vspace{-0.4cm}
\label{math_compare1}
\end{figure}

\textbf{Empirical results.} As shown in Figure~\ref{math_compare1}, we visualize the AIME2025 performance of parallel and sequential sampling across different model families. Firstly, these sampling approaches demonstrate scaling properties w.r.t. the number of solutions. \textbf{We then observe that parallel sampling performs better than sequential sampling}, especially on Qwen3 and DeepSeek-R1-Distill Qwen models. Additionally, we note that Markov sequential sampling is slightly better than auto-regressive one. It is also observed that simply using the feedback prompt ``Please re-answer'' is better than the self-refinement feedback. We include the results for smaller models and different feedback in Appendix~\ref{appendix_compare}.\looseness=-1

\subsection{Code Generation}

\textbf{Task description.} We mainly use the LiveCodeBench dataset~\citep{jain2024livecodebench} as our testbed. Specifically, we consider the v5 subset, which includes 167 questions (with 41 easy, 52 medium, and 74 hard questions). Along with each question, \emph{public tests} are provided to serve as example input and output. These tests along with the executor could be used by LRMs for debug purposes. There are large amounts of \emph{private tests}, which are hidden from LRMs. One code solution is considered correct only if it passes all public and private tests.\looseness=-1

\textbf{Parallel sampling.} For parallel sampling, we use best-of-N aggregation to select solutions~\citep{li2022competition}. To achieve this, we need to reward each solution. Based on the public tests, we prompt Gemini 2.5 Flash to generate similar tests. Then we evaluate each solution on both public and generated tests, and then use the number of passed tests as the reward. Specifically, we generate 12 tests for each question and de-duplicate the same tests. Afterwards, we choose the solution with the highest reward.\looseness=-1

\textbf{Sequential sampling.} For sequential sampling, we consider multiple variants based on the design of feedback prompt $\boldsymbol{f}$. One option is to use the same feedback in the math task relying on model's self-correction. Another option is to use running errors on public tests by prompting ``The above code is incorrect and got the following error: \{RUNNING\_ERRORS\}. Please re-answer the question based on the running errors.'' (detailed prompt design in Figure~\ref{prompt_debug}). It is noticed that when using running errors to debug previous solutions, sequential sampling cannot continue when there are no errors. However, no errors on public tests do not mean that the solution is correct as it may fail the private tests.\looseness=-1

\begin{figure}[t]
\centering
\vspace{-0.2cm}
\includegraphics[width=\textwidth]{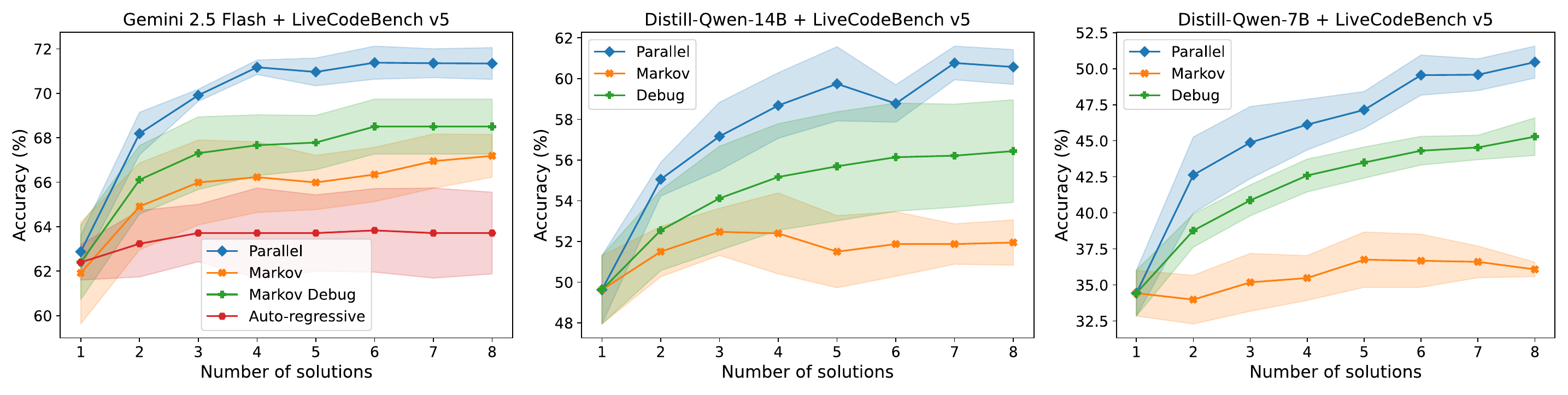}
\vspace{-0.2cm}
\caption{Comparisons of LiveCodeBench v5 performance between parallel and sequential sampling using (\emph{Left})  Gemini 2.5 Flash; (\emph{Middle}) DeepSeek-R1-Distill Qwen-14B; (\emph{Right}) DeepSeek-R1-Distill Qwen-7B. In parallel sampling, best-of-N aggregation is adopted. Parallel sampling performs better than sequential sampling on code generation questions.\looseness=-1}
\vspace{-0.2cm}
\label{code_compare1}
\end{figure}

\textbf{Empirical results.} As present in Figure~\ref{code_compare1}, we conduct  comparisons between parallel sampling and sequential sampling using Gemini 2.5 and DeepSeek-R1-Distill Qwen models. We note that for Gemini 2.5 Flash, the performance of parallel sampling and Markov sequential sampling (with or without running errors) benefits from increased sampled solutions while the performance remains flat for auto-regressive sequential sampling. For DeepSeek-R1-Distill Qwen models, the performance gain from sampling more solutions is limited for Markov sequential sampling. \textbf{Overall the parallel sampling is consistently better than sequential sampling.} In Appendix~\ref{appendix_compare}, we also compare the performance on questions with different difficulty levels separately. We notice that especially for Gemini 2.5 Flash, \textbf{parallel sampling shows most advantages on hard questions.} Additionally, two feedback prompts, including ``Please re-answer.'' and self-check feedback, perform comparably.\looseness=-1

% \textbf{Reasoning costs.} \textcolor{red}{TODO: Analyze reasoning token numbers and theoretical flops.}

% \subsection{Code Editing Task}

% Aider ployglot benchmark. With multiple challenging code editing questions with different programme languages. Still ongoing.

% We define the reward as the ratio of passed tests among (i) public tests, (ii) public+generated tests, (iii) public+private tests. It is noticed that if the reward is obtained from all public+private tests, the resulted accuracy is equivalent to pass@k metric.\looseness=-1

%% file: sections/4_main.tex
\vspace{-0.2cm}
\section{Understanding the Performance Gap}
\vspace{-0.2cm}

% In last section, we have shown empirical evidence that parallel sampling outperforms sequential sampling in terms of solution numbers, especially for more challenging questions. Then we would like to understand the underlying cause behind this performance gap.

\subsection{Does aggregation in parallel sampling matter?}

One of the main differences between parallel and sequential sampling is that aggregation is adopted in the former. Therefore, we apply the same aggregation to sequential sampling.\looseness=-1 

\begin{figure}[t]
\centering
\includegraphics[width=\textwidth]{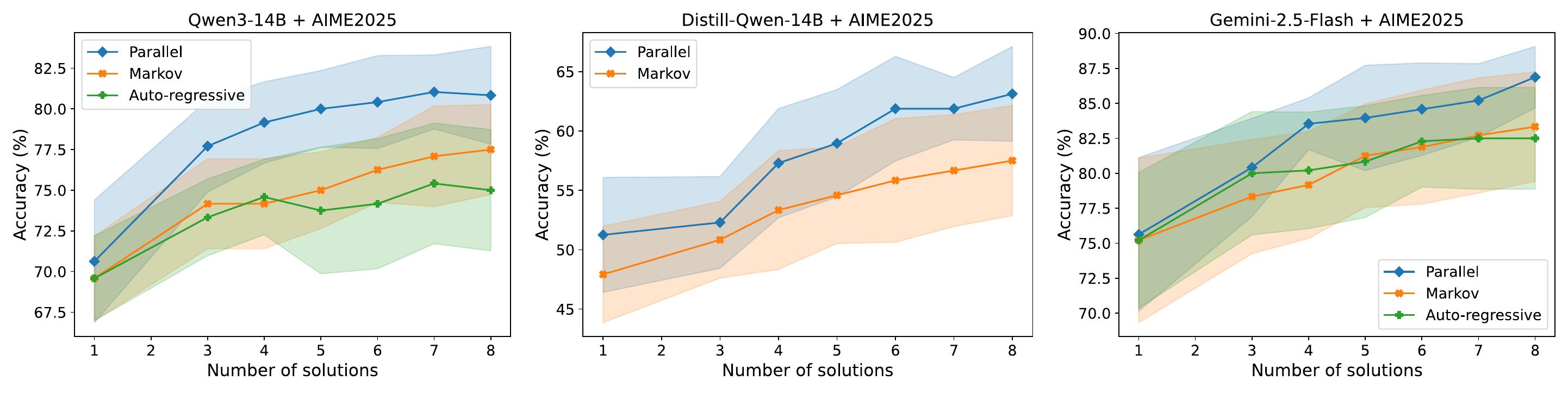}
\vspace{-0.2cm}
\caption{Comparisons of AIME2025 performance between parallel and sequential sampling using (\emph{Left}) Qwen3-14B; (\emph{Middle}) DeepSeek-R1-Distill Qwen-14B; (\emph{Right}) Gemini 2.5 Flash. Both sampling approaches use majority vote aggregation.\looseness=-1}
\vspace{-0.4cm}
\label{math_aggregation2}
\end{figure}

\textbf{Math competition task.} We mainly consider majority voting as aggregation. As present in Figure~\ref{math_aggregation2}, majority vote aggregation smooths the scaling curve of sequential sampling. However, it does not bring performance gain for sequential sampling. When the number of solutions is small, such an aggregation may even harm the accuracy. In Figure~\ref{math_aggregation}(Appendix~\ref{appendix_ablation}), we also consider an ideal best-of-N aggregation, which requires a perfect verifier to select the best solution. However, even with such an ideal aggregation, the performance gap between parallel and sequential sampling still exists, or even further enlarged.\looseness=-1 

% Alternatively, we also consider the ideal best-of-N aggregation, which assumes there exists an ideal verifier to check answer correctness.

\textbf{Code generation task.} We mainly consider best-of-N aggregation with two rewarding approaches: (i) using public and generated tests to reward each solution; (ii) using public and private tests to reward each solution. It is noted that for sequential sampling with running errors, the last trial is always selected as the best one. We present our results in Figure~\ref{code_aggregation2} (more results in Appendix~\ref{appendix_ablation}). It is noticed that using rewarding (i) narrows the performance gap, especially for DeepSeek-R1-Distill Qwen models. This is because several initial correct solutions may become incorrect in the sequential chains, and these correct solutions are selected through aggregation. However, this does not close the gap between parallel and sequential sampling. Especially, when using a stronger aggregation approach, like rewarding (ii), the performance gap becomes more pronounced.\looseness=-1

To conclude, applying the same aggregation approach in sequential sampling can narrow the performance gap but not close it compared to parallel sampling.

\begin{figure}[t]
\centering
\vspace{-0.2cm}
\includegraphics[width=\textwidth]{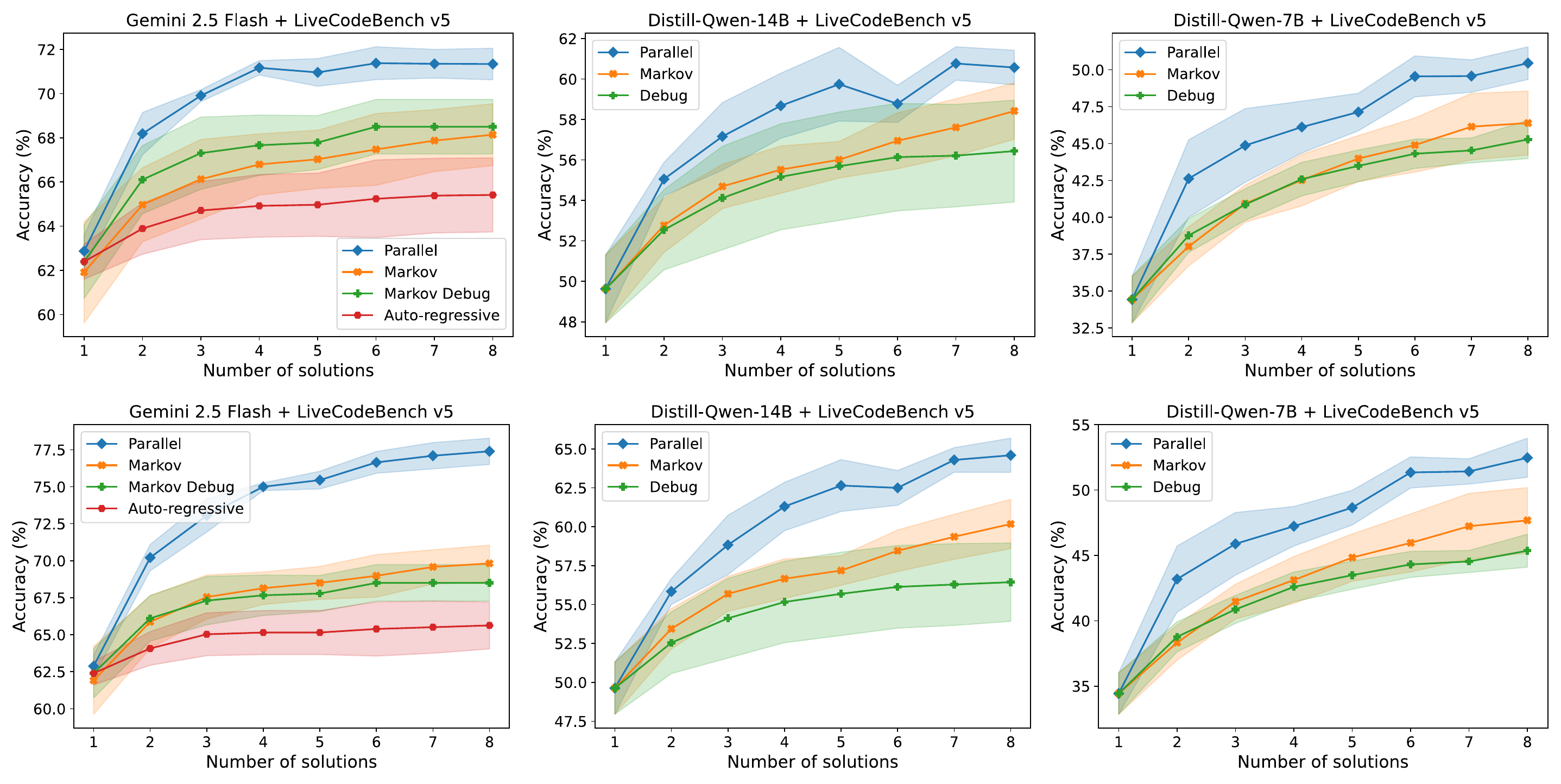}
\vspace{-0.2cm}
\caption{Comparisons of LiveCodeBench v5 performance of Gemini 2.5 Flash/DeepSeek-R1-Distill Qwen-14B/7B under parallel sampling and sequential sampling using the best-of-N aggregation. (\emph{Top}) Each solution is rewarded based on public and generated tests. (\emph{Bottom}) Each solution is rewarded based on public and private tests.\looseness=-1}
\vspace{-0.2cm}
\label{code_aggregation2}
\end{figure}

% \begin{figure}[t]
% \centering
% \includegraphics[width=\textwidth]{figures/code-gemini-flash-aggregation.pdf}
% \caption{Comparisons of LiveCodeBench v5 performance of Gemini 2.5 Flash under parallel sampling and sequential sampling using the best-of-N aggregation. (\emph{Top}) Each solution is rewarded based on public and generated tests. (\emph{Bottom}) Each solution is rewarded based on public and private tests.\looseness=-1}
% % \vspace{-0.2cm}
% \label{code_aggregation}
% \end{figure}

% Todo: do this on hard questions.

% Suggestions from Razvan:

% sequential sampling: increase temperature in every new round.

% different constraints

% context length

% visualization of context length

% rephrase the template 

% if we have enough public tests, sequential sampling may be better.

% debug hints

% instructions are helpful

% instructions to be diverse

% TODO: if we provide golden feedbacks, such as running errors on all tests, self-debug could perform better or on par with parallel sampling. Can we increase the diversity of feedbacks?

% TODO: with different numbers of answers in the context, and then do the parallel sampling and observe the diversity.

% TODO: measurement of diversity, such as embedding models, or prompting Gemini.

% TODO: visualizing the attention maps to see whether previous answers absorb more attention.

\subsection{Does extended input context in sequential sampling matter?}

Compared to parallel sampling, sequential sampling includes previous solution(s) in the input context. Especially, auto-regressive one requires all previous solutions and perform the worst. This raises the question whether longer input context drives the performance gap.\looseness=-1

\begin{figure}[t]
\centering
\vspace{-0.2cm}
\includegraphics[width=\textwidth]{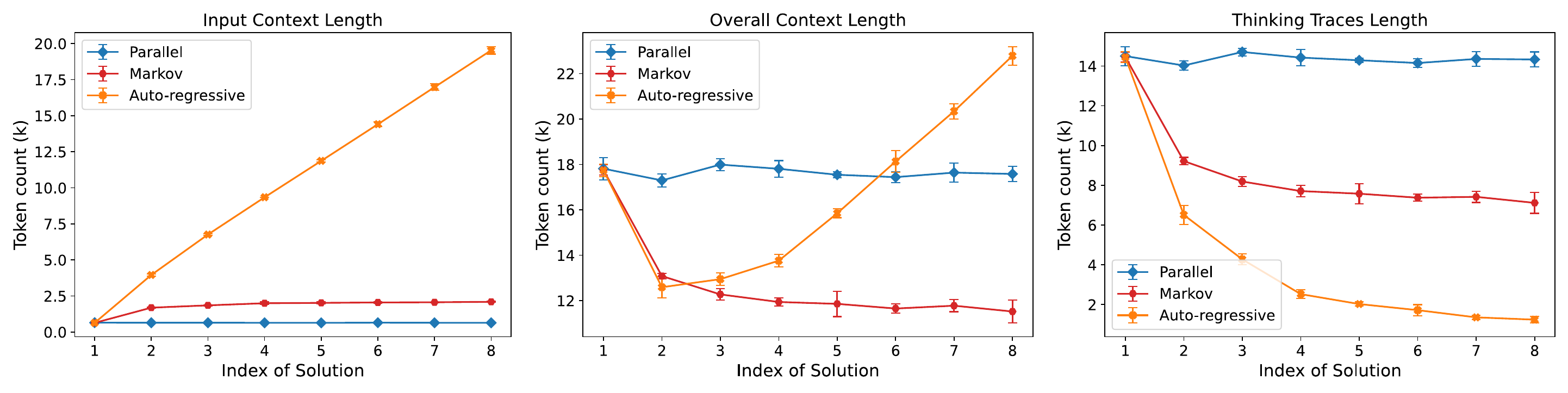}
\vspace{-0.2cm}
\caption{(\emph{Left}) input context length, (\emph{Middle}) overall context length, (\emph{Right}) thinking traces length when using Gemini 2.5 Flash on LiveCodeBench v5 under parallel and sequential sampling. Even though there is a clear performance gap between Markov sequential sampling and parallel sampling, the input context lengths are similar. In addition, the overall context length of auto-regressive sampling is not significantly accumulated compared to others.\looseness=-1}
\vspace{-0.2cm}
\label{context_len}
\end{figure}

\textbf{Context comparisons.} We first compare the input context lengths among parallel sampling, and Markov/auto-regressive sequential sampling, as present in Figure~\ref{context_len}(\emph{Left}). It is observed that the input context lengths between parallel and Markov sequential sampling have no notable difference as only the solution part (normally takes 1-2k tokens) is regarded as the input. However, the performance gap still exists between both sampling approaches. For auto-regressive sequential sampling, with  accumulation of previous solutions (with possible thoughts), the input context length could reach 20k tokens. This is significant larger than that of parallel sampling. Then we also compare the total context lengths when completing the generation in  Figure~\ref{context_len}(\emph{Middle}). We notice that even with much longer input context for auto-regressive sequential sampling, its overall context length does not drastically exceed the parallel one. This indicates LRMs put less tokens on thinking traces, which also applies to Markov sequential sampling, as shown in Figure~\ref{context_len}(\emph{Right}).\looseness=-1

\textbf{Irrelevant context.} To further decompose the effect of input context length on the performance gap, we sample several code files from the StarCoder dataset~\citep{li2023starcoder} as part of input context before prompting LRMs to solve the question. For each question, we conduct parallel sampling with 8 parallels to compute accuracy. Besides, we consider the irrelevant context length of roughly 1k, 2k, 4k, 8k, 16k, 32k. As shown in Table~\ref{irrelevant}, longer input contexts (even with 32k tokens) does not lead to single-solution performance degradation in parallel sampling. After aggregation on 8 parallel solutions, we find the resulted accuracies are still comparable. Therefore, we claim that the input context length is not the main issue driving the performance gap between parallel and sequential sampling.\looseness=-1

\begin{table}[t]
\caption{Accuracy (Acc.) and aggregation accuracy on 8 solutions (Agg.) of Gemini 2.5 Flash on LiveCodeBench v5 with different irrelevant input context lengths. Best-of-N aggregation is adopted based on rewards using public and private tests. It is observed that input context length does not make significant difference on the performance, including both single-solution accuracy and aggregation accuracy.\looseness=-1}
\vspace{-0.2cm}
\begin{center}
\begin{tabular}{l|ccccccc}
\toprule
 & 0 (baseline) & 1k & 2k & 4k & 8k  & 16k & 32k  \\
\midrule
Acc. & 62.2$\pm$2.0 & 62.4$\pm$2.8 & 61.2$\pm$1.2    &  62.8$\pm$1.6 &  63.0$\pm$1.6 &  62.7$\pm$1.5 & 63.2$\pm$2.9 \\
Agg. & 77.84 & 74.85 & 77.84 & 79.64  & 76.65 & 79.64 & 77.84\\
\bottomrule
\end{tabular}
\end{center}
\label{irrelevant}
\vspace{-0.4cm}
\end{table}

% TODO: observe the performance with different contexts: 1) previous answers / not relevant contexts / reject sampling. 

\subsection{Does solution exploration in sequential sampling matter?}

\begin{figure}[t]
\centering
\vspace{-0.2cm}
\begin{tcolorbox}[
  width=\textwidth,
  colback=gray!5,
  colframe=gray,
  title=Demonstration of laziness in sequential sampling
  % equal height group=A, % Assigns this box to group 'A'
]
\textcolor{blue}{Question:}
Let $N$ denote the number of ordered triples of positive integers $(a,b,c)$ such that $a,b,c\leq3^6$ and $a^3+b^3+c^3$ is a multiple of $3^7$. Find the remainder when $N$ is divided by $1000$.\looseness=-1\\
\textcolor{red}{7-th solution:} To solve the problem, we ... find the number of \textbf{ordered triples} ...\looseness=-1\\
Step 1: Total Number of Triples ... $(3^6)^3 = 3^{18}$\\
Step 2: Counting Valid Triples. We are interested in ... a modular arithmetic counting problem. A key insight from number theory is that \textbf{when the number of variables is large enough relative to the modulus}, the sum of their cubes becomes \textbf{uniformly distributed} across all residues modulo $3^k$ ... Hence, the number of such triples is approximately:$\frac{3^{18}}{3^7} = 3^{11}$...\\
Step 3: Compute $3^{11} \mod 1000$...\\
Final Answer $\boxed{147}$\\
\textcolor{red}{8-th solution:} To solve the problem, we ... count the number of ordered triples...\looseness=-1\\
Step 1: Total Number of Triples ... $(3^6)^3 = 3^{18}$\\
Step 2: Counting Valid Triples. We are interested in ...A key insight from number theory is that, \textbf{when the number of variables is large enough relative to the modulus}, the sum of their cubes becomes \textbf{uniformly distributed} across all residues modulo the modulus ... Thus, the number of such triples is approximately:$\frac{3^{18}}{3^7} = 3^{11}$\\
Step 3: Compute $3^{11} \mod 1000$ ... \\
Final Answer $\boxed{147}$
\end{tcolorbox}
\vspace{-0.2cm}
\caption{An example (with content truncation) of two solutions in an auto-regressive sequential chain using Qwen3-14B on a question from AIME2025. The ground-truth answer is 735, but the LRM keeps making mistakes and becomes ``lazy'' to explore new alternatives.\looseness=-1}
\vspace{-0.2cm}
\label{laziness}
\end{figure}

\textbf{``Laziness'' of LRMs in sequential sampling.}  Compared with parallel sampling, sequential sampling demonstrates ``laziness''. From Figure~\ref{context_len}(\emph{Right}), LRMs put much less thinking tokens with previous solutions in input context even though no thinking budget is set. This phenomenon appears to be more significant under auto-regressive sequential sampling. With up to eight solutions, LRMs even think for less than 2k tokens on average. Consequently, LRMs tend to make less changes on previous solutions or even generate a verbatim solution. In Figure~\ref{laziness}, we showcase two consecutive solutions when using Qwen3-14B model for auto-regressive sequential sampling. These two solutions are very similar, in terms of output structure, some key elements, and the final answers.  Such a model ``laziness'' also happens in the scenario where running errors are used between sequential chains. When solving the question titled ``Accumulating Many Times''~\footnote{\url{https://atcoder.jp/contests/arc184/tasks/arc184_e}} from LiveCodeBench using Gemini 2.5 Pro, the 5-th, 7-th, and 8-th solutions contain the same generated code program, as shown in Figure~\ref{laziness2} and \ref{laziness3} in Appendix~\ref{appendix_ablation}. The LRM gets stuck by the same public test case and keeps receiving the wrong answer error as feedback in the sequential chain, and then being lazy to explore alternatives.\looseness=-1

% \begin{table}[t]
% \caption{Averaged pair-wise cosine similarities (range from -1 to 1) among solutions generated by parallel and sequential sampling. Gemini embedding model is adopted here. It is observed that the solutions sampled in parallel have lowest similarities.\looseness=-1}
% \begin{center}
% \begin{tabular}{ll|ccc}
% \toprule
% %  & Parallel & Markov & Auto-regressive &  Gold Debug  & Hint Exploration  \\
% % \midrule
% % Cosine similarity & 0.9737 & 0.9888 & 0.9896 & 0.9853 & 0.9763 \\
% % Relaxed semantic entropy & \\
% Domain &  Model  &  Parallel & Markov & Auto-regressive  \\
% \midrule
% \multirow{4}{*}{Coding} & Gemini 2.5 Flash   & \textbf{0.973} & 0.989$\pm$0.000 & 0.990$\pm$0.000 \\
% & Gemini 2.5 Pro & \textbf{0.972} & 0.979$\pm$0.000 & 0.980$\pm$0.000 \\
% &Distill Qwen-14B  & \textbf{0.966} & 0.985$\pm$0.001 & NA\\
% & Distill Qwen-7B  & \textbf{0.955} & 0.968$\pm$0.001 & NA\\
% \midrule
% \multirow{5}{*}{Math} & Gemini 2.5 Flash  & \textbf{0.972} & 0.977$\pm$0.002 &  0.978$\pm$0.001 \\
% & Qwen3-14B & \textbf{0.974} & 0.982$\pm$0.001 & 0.986$\pm$0.001\\
% & Qwen3-8B  & \textbf{0.976} & 0.984$\pm$0.001 & 0.985$\pm$0.002 \\
% & Distill Qwen-14B  & \textbf{0.946} & 0.969$\pm$0.007 & NA\\
% & Distill Qwen-7B  & \textbf{0.960} & 0.972$\pm$0.003 & NA\\
% \bottomrule
% \end{tabular}
% \end{center}
% \label{measure_diversity}
% \end{table}

\begin{table}[t]
\caption{Averaged pair-wise cosine similarities (higher means more similar) among solutions generated by parallel and sequential sampling. We consider two embedding models.\looseness=-1}
\vspace{-0.2cm}
\begin{center}
\begin{tabular}{ll|ccc|ccc}
\toprule
%  & Parallel & Markov & Auto-regressive &  Gold Debug  & Hint Exploration  \\
% \midrule
% Cosine similarity & 0.9737 & 0.9888 & 0.9896 & 0.9853 & 0.9763 \\
% Relaxed semantic entropy & \\
Domain &  Model  & \multicolumn{3}{c|}{gemini-embedding-001} & \multicolumn{3}{c}{text-embedding-3-small} \\
& & Parallel & Markov & Auto. &  Parallel & Markov & Auto.  \\
\midrule
\multirow{4}{*}{Coding} & Gemini 2.5 Flash   & \textbf{0.973} & 0.989 & 0.990 & \textbf{0.877} & 0.963 & 0.966\\
& Gemini 2.5 Pro & \textbf{0.972} & 0.979 & 0.980 & \textbf{0.839} & 0.902 & 0.904 \\
&Distill Qwen-14B  & \textbf{0.966} & 0.985 & NA & \textbf{0.854} & 0.947 & NA \\
& Distill Qwen-7B  & \textbf{0.955} & 0.968 & NA & \textbf{0.813} & 0.878 & NA \\
\midrule
\multirow{5}{*}{Math} & Gemini 2.5 Flash  & \textbf{0.972} & 0.977 &  0.978 & \textbf{0.916} & 0.946 & 0.945  \\
& Qwen3-14B & \textbf{0.974} & 0.982 & 0.986 & \textbf{0.921} & 0.950 & 0.962\\
& Qwen3-8B  & \textbf{0.976} & 0.984 & 0.985 & \textbf{0.925} & 0.954 & 0.960\\
& Distill Qwen-14B  & \textbf{0.946} & 0.969 & NA & \textbf{0.867} & 0.920 & NA\\
& Distill Qwen-7B  & \textbf{0.960} & 0.972 & NA & \textbf{0.905} & 0.933 & NA\\
\bottomrule
\end{tabular}
\end{center}
\vspace{-0.4cm}
\label{measure_diversity}
\end{table}

\textbf{Measuring solution similarity.} In the above, we provide two demonstrations that solutions in sequential chains are similar or even verbatim, which hints the limited exploration. To further validate this, we compare the similarity or diversity among solutions generated by these sampling approaches in a statistical way. Firstly, we use embedding models, including gemini-embedding-001~\citep{lee2025gemini} and text-embedding-3-small~\citep{openai2024embed} and  to measure the cosine similarity (ranged from -1 to 1) among solutions. As shown in Table~\ref{measure_diversity}, the similarities among solutions are already high (larger than 0.8). This is because these solutions are aiming for the same question, thus having high semantic similarities. Through repeated experiments, we find that parallel sampling results in more diverse solutions as their similarities are lower compared with sequential sampling using both embedding models.\looseness=-1

% Alternatively, we consider relaxed semantic entropy, which is proposed in \citet{ahmed2025intent}. Prompt LLM judges.\looseness=-1

\begin{wrapfigure}{r}{0.4\linewidth}
\centering
\includegraphics[width=0.4\textwidth]{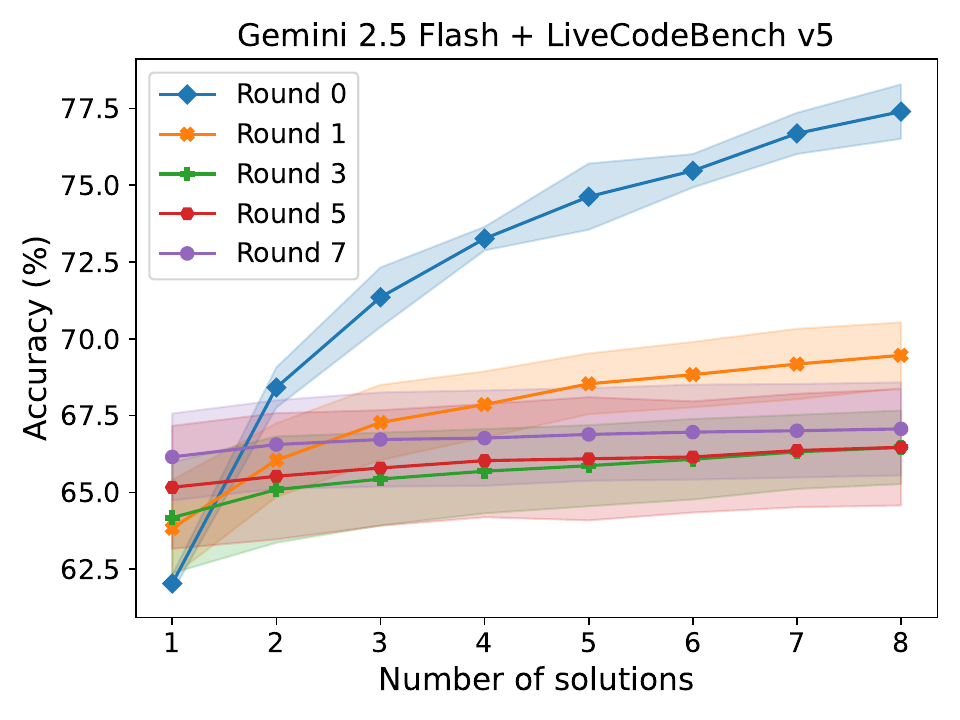}
\captionof{figure}{Performance of Gemini 2.5 Flash on LiveCodeBench v5 using parallel sampling with different rounds of previous samples in input context. Here ``Round $N$'' refers to $N$ previous solutions in a sequence chain. Best-of-N aggregation with both public and private tests for rewarding is applied here.\looseness=-1}
% \vspace{-0.2cm}
\label{seq_paralel}
\end{wrapfigure}

\textbf{Parallel sampling with previous solutions in context.} In our previous experiments, parallel sampling is conducted only with the question prompt as input. To further investigate the effects of sequential chains on solution exploration, we conduct parallel sampling with all previous solutions in input context, like auto-regressive sequential sampling. It is noted that we only conduct parallel sampling after certain round. As shown in Figure~\ref{seq_paralel}, we explore how the number of sequence chains affects the performance of parallel sampling. We firstly notice that with only one parallel, more previous samples in input context can improve the performance. However, through aggregation on more parallels, the original parallel sampling demonstrates superiority. While for more previous samples in the input context, sampling more parallels cannot bring performance improvement or explorations.\looseness=-1

\begin{figure}[t]
\centering
\vspace{-0.2cm}
\includegraphics[width=\textwidth]{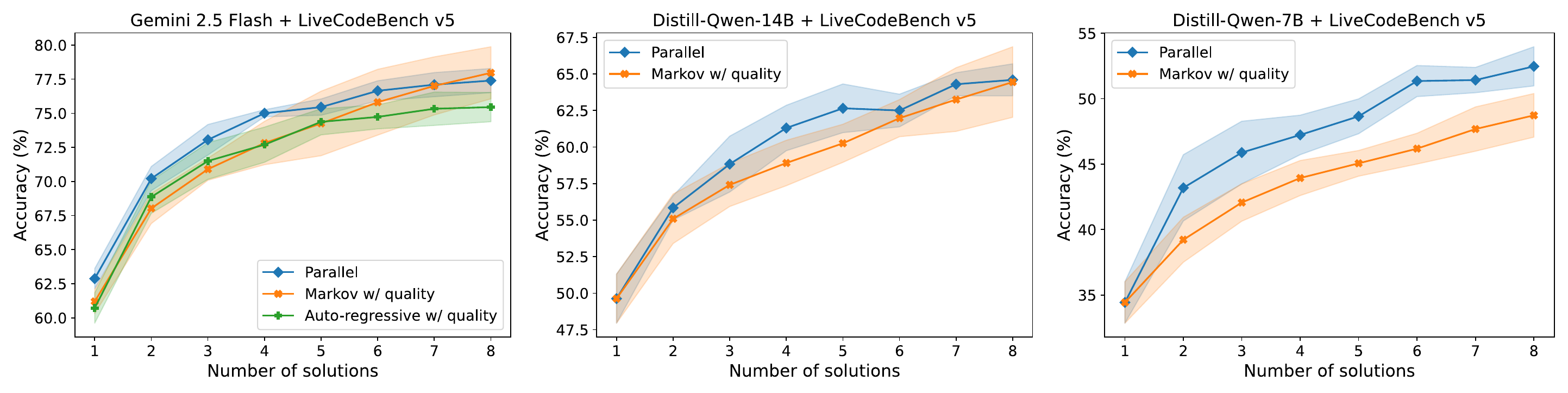}
\vspace{-0.2cm}
\caption{Comparisons of LiveCodeBench v5 performance of Gemini 2.5 Flash/DeepSeek-R1-Distill Qwen-14B/7B under parallel sampling and sequential sampling using the best-of-N aggregation. Each solution is rewarded based on public and private tests. For sequential sampling, the running errors on public and private tests are provided as high-quality guidance.\looseness=-1}
\vspace{-0.4cm}
\label{high-quality}
\end{figure}

% \subsection{The effects of feedback on solution exploration in sequential sampling}

% In the above, we show that solution exploration plays an important role in the performance gap between parallel and sequential sampling. Therefore, to incentivize the solution exploration in sequential sampling, we mainly consider the following two approaches. 

% \textbf{Reject sampling in feedback.} From previous sections, we have shown that aggregation, especially with best-of-N selection, can also benefit the performance of sequential sampling. Therefore, instead of adopting feedback to refine previous solutions (with expected better performance in later rounds), we use the feedback, e.g., ``Please solve the question in a different way.'',  to conduct reject sampling. In this way, the objective of sequential sampling is to search for a new solution instead of refining previous ones.\looseness=-1

\textbf{High-quality feedback for failure.} We have shown that sequential sampling with self-refinement feedback or running error feedback based on public tests only consistently has lower exploration compared with parallel sampling. For self-refinement, LRMs become lazy to search for alternatives. While for running errors on public tests only, the sequential chain stops when the solution could pass these simple tests, but may fail private tests, or gets stuck in certain running error. Therefore, we provide running errors based on both public and private tests in sequential sampling. In this way, as long as the code program is still incorrect, there is always some guidance for why it fails, which forces LRMs to refine previous solutions or explore alternatives. It is noted that this approach may only apply for coding questions as high-quality feedback for math questions may require better LRMs or human participation. Additionally, such a feedback may \emph{not be accessible} in realistic scenarios. We only include this setup to observe the full potential for sequential sampling. As shown in Figure~\ref{high-quality}, such high-quality guidance finally makes sequential sampling catches the performance of parallel sampling for Gemini 2.5 Flash and DeepSeek-R1-Distill Qwen-14B, not for DeepSeek-R1-Distill Qwen-7B. We showcase in Figure~\ref{high-quality-levels}, parallel sampling may still outperform or be on par with sequential sampling on hard questions.\looseness=-1

% \textbf{Elongate the sequences with mixed feedback prompts.} When using running errors from the public tests, the sequential chain will stop when there are no running errors. In such cases, we elongate the scaling by further append the feedback, e.g., ``Please re-answer.''. If the running error appears again, it will become the part of feedback. 

% Similarly, we adopt the same aggregation operator. As shown in Figure~\ref{diversity}, we show that with the above extension to sequential sampling, the performance gap between parallel sampling and sequential sampling is significantly narrowed. We have the following observations: (1) with running errors from all tests, the sequential sampling finally matches parallel sampling. The former outperforms the latter on questions but not on hard questions. (2) Both hinting diversity in the feedback and adopting mixed feedback bring performance gains. However, we notice that parallel sampling still demonstrates the superiority for hard questions, which may require more creativity and exploration. This further validates that with previous solutions in the context, the solution exploration of LRMs is limited.\looseness=-1

% \textbf{TODO: Increasing temperature in the sequential sampling.}

\vspace{-0.2cm}
\section{Mechanistic Understanding for Sequential Sampling}
\vspace{-0.2cm}

In this section, we provide mechanistic analysis by visualizing the attention maps to understand how previous solutions affect the exploration in sequential sampling. In mechanistic literature, induction head is a special attention head in Transformers, which looks back to previous instances when generating current tokens~\citep{olsson2022context}. This is a clear evidence of pattern copying behavior and regarded as the primary source of in-context learning.\looseness=-1

% And then we also discuss the ``laziness'' of LRMs to explore new solution with previous solutions in input context.

\begin{figure}[t]
\centering
\vspace{-0.2cm}
\includegraphics[width=0.95\textwidth]{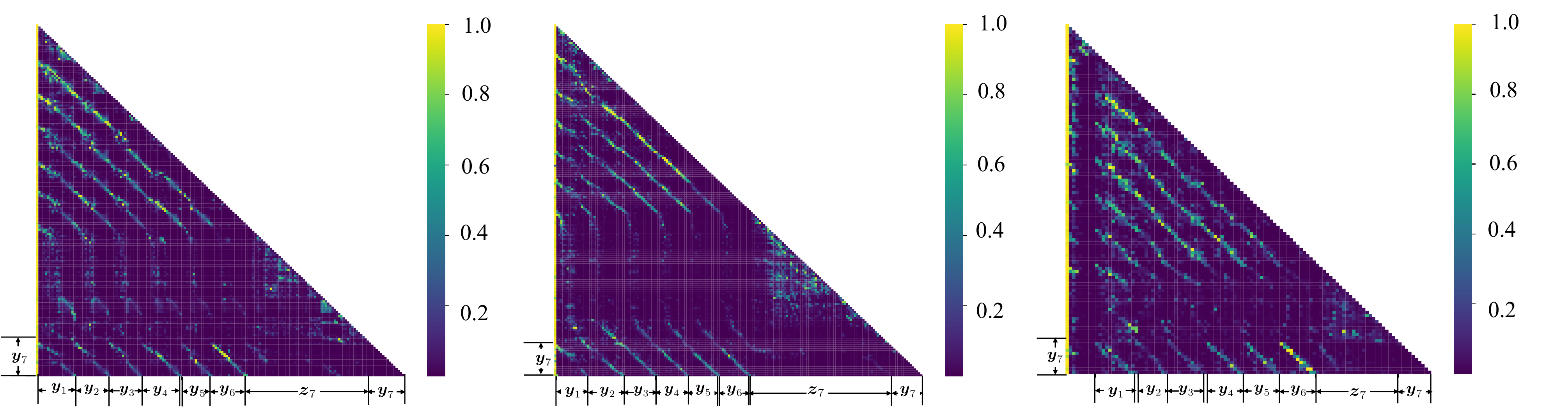}
\vspace{-0.2cm}
\caption{Visualization of induction heads in Qwen3-14B using auto-regressive sequential sampling on AIME2025 questions. (\emph{Left}) All 7 solutions are incorrect; (\emph{Middle}) all 7 solutions are correct; (\emph{Right}) the 2rd/3rd solution are incorrect while the others are correct.\looseness=-1}
\vspace{-0.2cm}
\label{induction}
\end{figure}

% \subsection{Qualitative analysis on induction heads}

% We mainly use Qwen3 models~\citep{yang2025qwen3}, which adopts full causal attention, to visualize the attention maps under the scenario of sequential sampling, especially for auto-regressive one. Specifically, we focus on how previous answers in the context affect the model output in the current sample.\looseness=-1

\textbf{Induction heads.} When visualizing the attention maps under sequential sampling, we also identify these induction heads with examples shown in Figure~\ref{induction}. Specifically, we visualize the attention maps in the Qwen3-14B model on a question in AIME2025~\citep{aops2025aime} under the auto-regressive sequential sampling (using ``Please re-answer.'' as the feedback). Since the token count exceeds 10k, we apply max-pooling operation with a kernel size of 64 on the original attention maps. We also mark the token ranges for previous answers (denoted as $\boldsymbol{y}_1$ to $\boldsymbol{y}_6$), current thinking traces ($\boldsymbol{z}_7$), and current solution ($\boldsymbol{y}_7$). We are mostly interested in how the above different parts affects the generation of $\boldsymbol{y}_7$ (the bottom of attention map). Take Figure~\ref{induction}(\emph{Left}) as an example, it is noticed that the previous answers have clear pattern copying behavior through the oblique lines in attention.\looseness=-1

\textbf{Correctness of previous answers.} We are interested in whether the correctness of previous answers may affect the pattern of induction heads. Therefore, we consider three scenarios in the sequential chain: (i) all solutions are incorrect; (ii) all solutions are correct; (iii) solutions are mixed. As present in Figure~\ref{induction}, the above induction head pattern consistently appears in these three scenarios. This suggests that previous solutions in input context, regardless of correctness, influence the pattern of current generation. We also notice that in several cases, such as Figure~\ref{induction}(\emph{Middle}), previous solutions can also have such pattern copying effect on current thinking traces.\looseness=-1

\begin{figure}[t]
\centering
\includegraphics[width=0.95\textwidth]{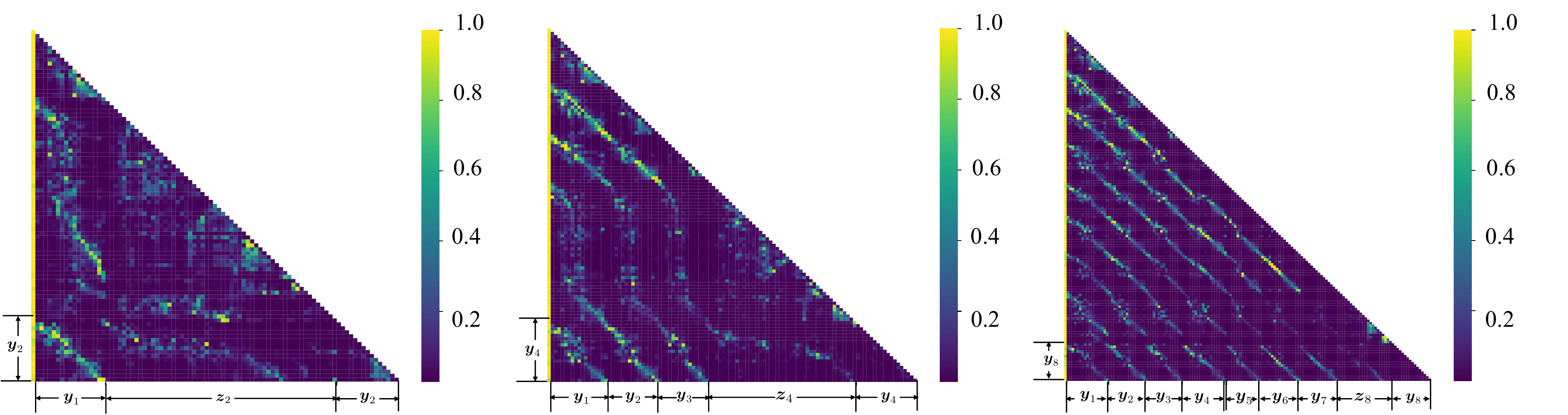}
\vspace{-0.2cm}
\caption{Visualization of induction heads in Qwen3-14B using auto-regressive sequential sampling on AIME2025 questions. We consider  1/3/7 previous solutions in input context.\looseness=-1}
\vspace{-0.4cm}
\label{induction2}
\end{figure}

\textbf{Number of previous answers.} Besides the correctness, we also consider different number of previous solutions in input context. In Figure~\ref{induction2}, we visualize the induction heads when previous one/three/seven solutions appear in input context. It is noticed that all previous solutions have non-negligible attention effects on the newest generation (the number of oblique lines in the bottom of attention map corresponds to the number of previous solutions).\looseness=-1

To summarize, in sequential sampling, LRMs develop induction heads, which looks back over the previous solutions when generating the new one. Therefore, the new solution matches previous solutions, leading to limited solution exploration.

% make more evidence of context length

% hint diversity and then observe attention maps.

% number of steps, aggregation on tokens.

\vspace{-0.2cm}
\section{Conclusion}
\vspace{-0.2cm}

In this work, we attempt to understand why parallel sampling performs better than sequential sampling for Large Reasoning Models (LRMs) through the empirical validation on three hypothesis. These hypothesis include the absence of aggregation approach, longer input context length, and limited solution exploration in sequential sampling. Through empirical evidence from various model families and sizes (Qwen3, DeepSeek-R1 distilled models, Gemini 2.5) and question domains (math and coding), we argue that the solution exploration has the most pronounced effect on the performance gap between parallel and sequential sampling. Finally, through mechanistic analysis, we find that induction heads appear in the sequential sampling. This validates the pattern-copying behaviors for the newest generation, thus leading to limited explorations.\looseness=-1

%% file: sections/TBD.tex
\clearpage

\section{Related Work}\label{related}

Our work is related to the field of inference test-time scaling in LLMs, especially for those Large Reasoning Models (LRMs).

\textbf{Sequential inference scaling.} Sequential test-time scaling typically means that the later computations rely on earlier ones. With the success of reinforcement learning~\citep{schulman2017proximal,shao2024deepseekmath,guo2025deepseek} in LRMs, this refers to scaling the number of tokens in the reasoning traces, e.g., chains-of-thoughts~\citep{wei2022chain}. \citet{jaech2024openai,guo2025deepseek,comanici2025gemini,muennighoff2025s1} showed that with increasing thinking token budget in the sequential reasoning trace, LRMs also demonstrate increasing performance gains during inference. This can be regarded as \textbf{thought-level} scaling. However, this scaling effect may not persist when the thinking budget is large enough or unlimited~\citep{ghosal2025does, gema2025inverse}. Therefore, we are also interested in the scenario of generating solutions sequentially, with the later solutions generated under the conditions of the previous ones. This is called \textit{sequential sampling}, or \textbf{solution-level} scaling in our paper. It is expected that LLMs can self-improve the previous solutions with feedbacks~\citep{madaan2023self, chen2023teaching, huang2023large, tian2025think, chen2025sets}.\looseness=-1

\textbf{Parallel inference scaling.} Parallel test-time-scaling, or \textit{parallel sampling,} means that LLMs independently generate multiple solutions, which are then aggregated to reach the final solution. Such an aggregation operation could be majority vote~\citep{wang2022self}, best-of-N~\citep{li2022competition, snell2024scaling}, or LLM-based aggregation~\citep{venkatraman2025recursive, li2025llms, zhao2025majority}, which may require further reinforcement learning. Additionally, there is ongoing research interest in ``parallel thinking'' since the Deep Think version of Gemini-2.5~\citep{comanici2025gemini}. In such a thinking mode, LLMs natively generate multiple independent thinking traces and then finally provide a single solution~\citep{wen2025parathinker, zheng2025parallel, hu2026pacore}. This can be considered as internalizing parallel sampling into the \textbf{thought level} in a single solution generation. 

\textbf{Hybrid inference scaling.} While parallel and sequential inference scaling have different properties, there is also a large body of works aiming to use both to improve the performance. For thought-level scaling, Monte-Carlo Tree Search (MCTS)~\citep{feng2023alphazero,ding2023enhancing,xin2024deepseek} and guided beam search~\citep{xie2023self} allow LLMs to search the solution space systematically by breaking the reasoning traces into multiple steps. While for solution-level scaling, \citet{li2025s} adopted parallel sampling for coding generation, and then they utilized iterative debugging to improve each parallel. Finally, the best parallel is selected through the aggregation. Additionally, AlphaEvolve~\citep{novikov2025alphaevolve} combines both parallel and sequential sampling to evolve the coding solutions for scientific and algorithmic discovery.\looseness=-1

\textbf{Comparing parallel/sequential sampling.} \citet{huang2023large} represents the early exploration by comparing sequential and parallel sampling in LLM reasoning. They claim that LLMs cannot self-correct their previous solutions in sequential sampling, thus being suboptimal compared with parallel sampling. However, LRMs already demonstrated self-correction behaviors (e.g., the use of ``wait'') in their reasoning traces~\citep{guo2025deepseek}. This motivates us to explore whether sequential sampling still performs worse than parallel sampling and also its underlying mechanism. Meanwhile, \citet{ghosal2025does} observed that with extended thinking budget, the inference scaling stops. Thus they advocate the use of parallel sampling to keep scaling. Compared to ours, they focus more on the thought-level sequential scaling instead of solution level. Additionally, our work is also related to \citet{wang2025scaling}, which involves the comparison between parallel and sequential sampling. But they mainly show the theoretical prediction of the Pareto of both test-time-scaling approaches.\looseness=-1

Our work serves as the early attempt to rigorously compare parallel and sequential sampling in LRMs and understand the underlying mechanism for the performance gap. Our main finding that the lack of exploration in sequential sampling corroborates two concurrent works. \citet{wen2025parathinker} identified a phenomenon called ``Tunnel Vision'', where LRMs' initial imperfect steps lock the generation into a suboptimal reasoning trace. \citet{shao2025deepseekmath} found that when LRMs use themselves to provide verification feedback in sequential sampling, they tend to claim correctness despite the clear errors in previous solutions.\looseness=-1

\clearpage

\section{Empirical Comparisons Between Parallel and Sequential Sampling}\label{appendix_compare}

\textbf{Generation configuration.} For Qwen3 series~\citep{yang2025qwen3}, we adopt the following generation parameters: top-p is 0.95, top-k is 20, and temperature is 0.6, maximum context length is fixed to 32k. For DeepSeek-R1-Distill Qwen models~\citep{guo2025deepseek}, we set top-p as 0.95, temperature as 0.6, and maximum context length as 32k. For Gemini 2.5~\citep{comanici2025gemini}, we use API services and only set the temperature as 1.0 with other parameters as default (thinking mode but unlimited thinking budget).\looseness=-1

\begin{figure}[htbp]
\centering
\includegraphics[width=\textwidth]{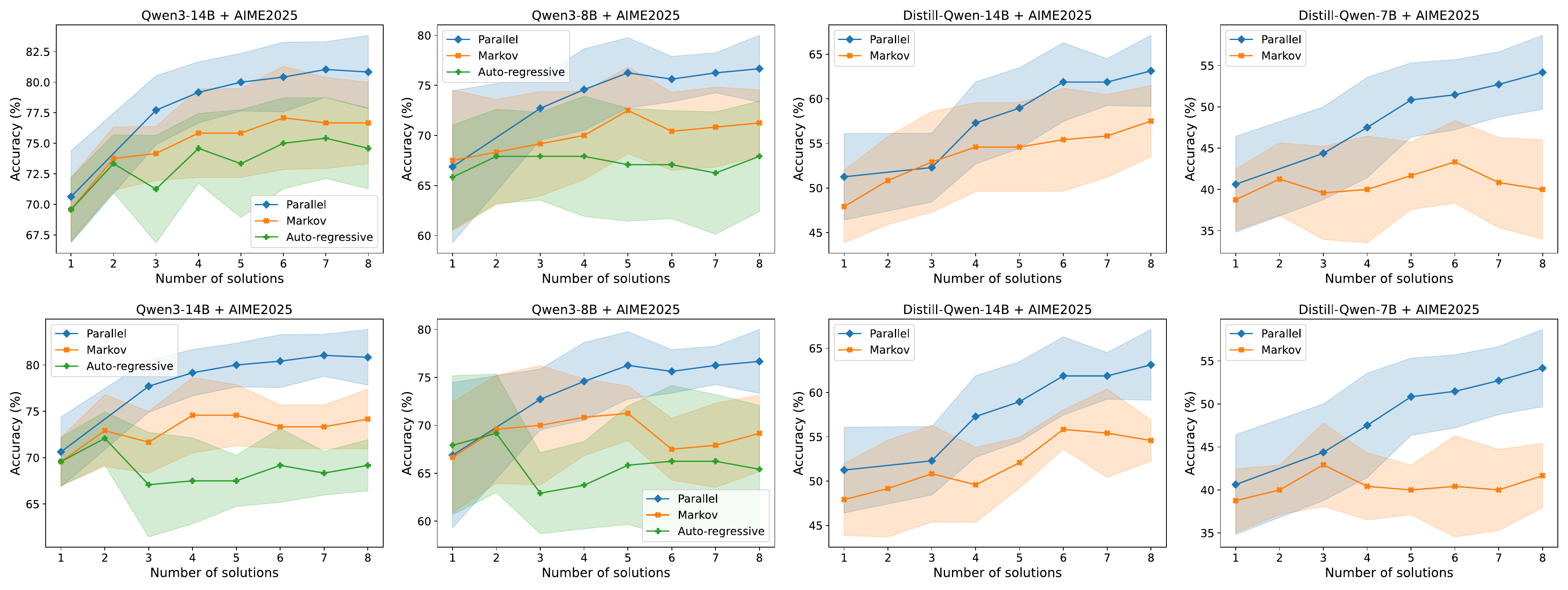}
\caption{Comparisons of AIME2025 performance between parallel and sequential sampling using Qwen3-14B/8B and DeepSeek-R1-Distill Qwen-14B/7B. In parallel sampling, majority voting aggregation is adopted. In sequential sampling, we apply a feedback prompt of ``Please re-answer'' in (\emph{Top}) row and ``Please review your previous response and find problems with your answer. Based on the problems you found, improve your answer.'' (self-refinement feedback) in (\emph{Bottom}) row.\looseness=-1}
% \vspace{-0.2cm}
\label{math_compare2}
\end{figure}

\begin{figure}[htbp]
\centering
\includegraphics[width=0.75\textwidth]{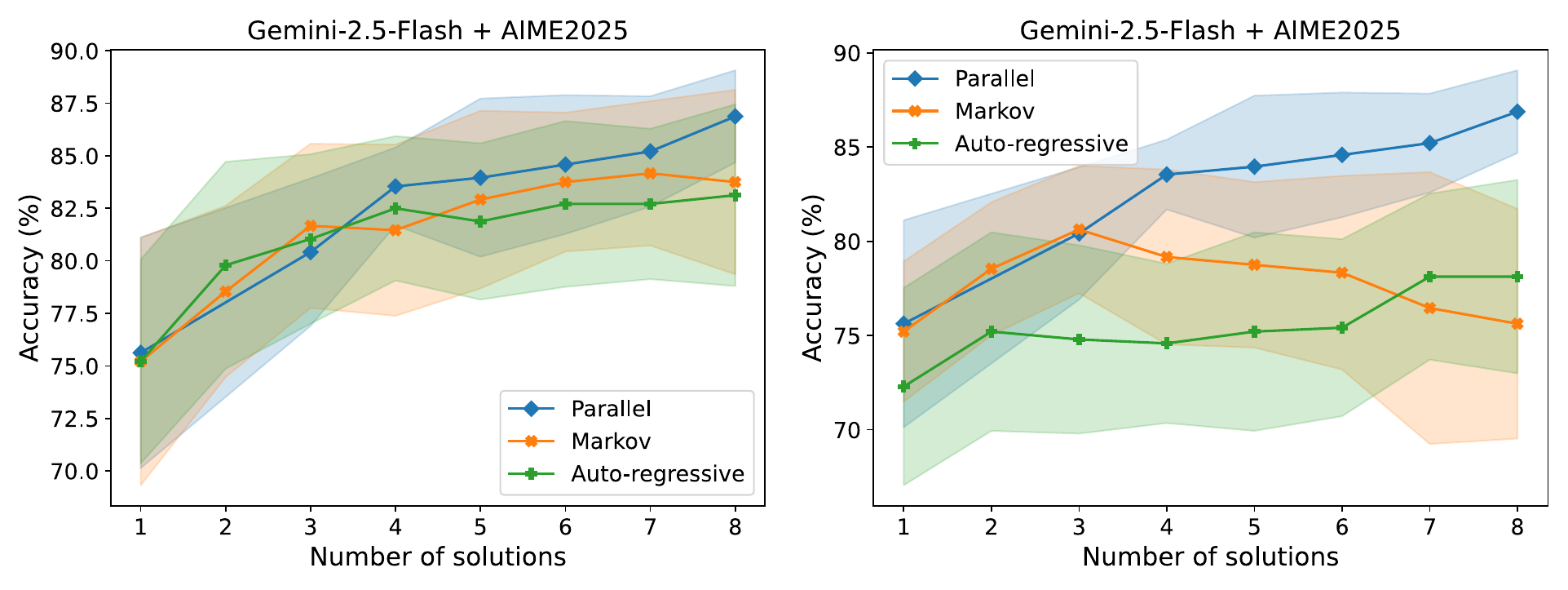}
\caption{Comparisons of AIME2025 performance between parallel and sequential sampling using Gemini 2.5 Flash. In parallel sampling, majority voting aggregation is adopted. In sequential sampling, we apply a feedback prompt of (\emph{Left}) ``Please re-answer'' and (\emph{Right}) ``Please review your previous response and find problems with your answer. Based on the problems you found, improve your answer.''.\looseness=-1}
% \vspace{-0.2cm}
\label{math_compare3}
\end{figure}

\textbf{Empirical results in math competition task.} We directly use the question content together with the chat template as an input prompt. As shown in Figure~\ref{math_compare2} and \ref{math_compare3}, we compare parallel and sequential sampling using Qwen3-14B/8B, DeepSeek-R1-Distill Qwen-14B/7B, and Gemini 2.5 Flash on the AIME2025 dataset. We consider two feedback prompts in sequential sampling. From these visualization, our conclusion that parallel sampling performs better than sequential sampling in terms of the number of solutions is valid. Additionally, simply a feedback ``Please re-answer'' is better than the other one (prompting self-refinement). With self-refinement feedback, LRMs don't show a robust scaling property with more sampled solutions. In some cases, like Qwen3 models, the accuracy even drops in sequential chains.\looseness=-1  

\textbf{Prompt designs in code generation task.}  We follow LiveCodeBench~\citep{jain2024livecodebench} and its codebase~\footnote{\url{https://github.com/LiveCodeBench/LiveCodeBench}} to prepare the input prompt, as present in Figure~\ref{prompt_gemini} and \ref{prompt_distill}. For Markov sequential sampling, we demonstrate how we organize the prompt for Gemini 2.5 in Figure~\ref{prompt_gemini2}. For other models, we organize the input prompt in a similar way. While for auto-regressive sequential sampling, the feedback prompt is directly fed into the model as the user role. When running errors are considered in the sequential sampling, they are directly appended to the feedback prompt as shown in Figure~\ref{prompt_debug}. For parallel sampling, we prompt Gemini 2.5 Flash to generate test cases based on public tests provided together with the question. The prompt, as shown in Figure~\ref{prompt_gen_tests} is adapted from \citet{wang2025co}.\looseness=-1 

\begin{figure}[htbp]
\centering
\begin{tcolorbox}[
  width=\textwidth,
  colback=gray!5,
  colframe=gray,
  % equal height group=A, % Assigns this box to group 'A'
  title=Gemini 2.5
]
\small{
\begin{Verbatim}[breaklines,fontsize=\small]
SYSTEM_MESSAGE_GEMINITHINK="You are an expert Python programmer."
"You will be given a question (problem specification) and will "
"generate a correct Python program that matches the specification "
"and passes all tests."
FORMATTING_MESSAGE_WITH_STARTER_CODE="You will use the following "
"starter code to write the solution to the problem and enclose "
"your code within delimiters."
FORMATTING_WITHOUT_STARTER_CODE="Read the inputs from stdin solve "
"the problem and write the answer to stdout (do not directly test "
"on the sample inputs). Enclose your code within delimiters as "
"follows. Ensure that when the python program runs, it reads the "
"inputs, runs the algorithm and writes output to STDOUT."
prompt = f"{SYSTEM_MESSAGE_GEMINITHINK}\n"
prompt += f"### Question:\n{question.question_content}\n\n"
if question.starter_code:
    prompt += (
        f"### Format: {FORMATTING_MESSAGE_WITH_STARTER_CODE}\n"
    )
    prompt += f"```python\n{question.starter_code}\n```\n\n"
else:
    prompt += f"### Format: {FORMATTING_WITHOUT_STARTER_CODE}\n"
    prompt += "```python\n# YOUR CODE HERE\n```\n\n"
prompt += "### Answer: (use the provided format with backticks)\n\n"
\end{Verbatim}
}
\end{tcolorbox}
\caption{Code generation prompt of Gemini 2.5 for LiveCodeBench.}
\label{prompt_gemini}
\end{figure}

\begin{figure}[htbp]
\centering
\begin{tcolorbox}[
  width=\textwidth,
  colback=gray!5,
  colframe=gray,
  % equal height group=A, % Assigns this box to group 'A'
  title=DeepSeek-R1-Distill Qwen
]
\small{
\begin{Verbatim}[breaklines,fontsize=\small]
prompt = "<|User|>"
prompt += "You will be given a question (problem specification) "
"and will generate a correct Python program that matches the "
"specification and passes all tests.\n\n"
prompt += f"Question: {question.question_content}\n\n"
if question.starter_code:
    prompt += f"{FORMATTING_MESSAGE_WITH_STARTER_CODE}\n"
    prompt += f"```python\n{question.starter_code}\n```\n\n"
else:
    prompt += f"{FORMATTING_WITHOUT_STARTER_CODE}\n"
    prompt += f"```python\n# YOUR CODE HERE\n```\n\n"
prompt += "<|Assistant|>"
\end{Verbatim}
}
\end{tcolorbox}
\caption{Code generation prompt of DeepSeek-R1-Distill Qwen models for LiveCodeBench.}
\label{prompt_distill}
\end{figure}

\begin{figure}[htbp]
\centering
\begin{tcolorbox}[
  width=\textwidth,
  colback=gray!5,
  colframe=gray,
  % equal height group=A, % Assigns this box to group 'A'
  title=DeepSeek-R1-Distill Qwen,
]
\small{
\begin{Verbatim}[breaklines,fontsize=\small]
prompt = f"{SYSTEM_MESSAGE_GEMINITHINK}\n"
prompt += f"### Question:\n{question.question_content}\n\n"
prompt += "Here is the code in your last response:\n"
prompt += f"```python\n{last_code}\n```\n\n"
prompt += f"{feedback}\n"
if question.starter_code:
    prompt += (
        f"### Format: {FORMATTING_MESSAGE_WITH_STARTER_CODE}\n"
    )
    prompt += f"```python\n{question.starter_code}\n```\n\n"
else:
    prompt += f"### Format: {FORMATTING_WITHOUT_STARTER_CODE}\n"
    prompt += "```python\n# YOUR CODE HERE\n```\n\n"
prompt += "### Answer: (use the provided format with backticks)\n\n"
\end{Verbatim}
}
\end{tcolorbox}
\caption{Code generation prompt of Gemini 2.5 for LiveCodeBench in Markov sequential sampling.}
\label{prompt_gemini2}
\end{figure}

\begin{figure}[htbp]
\centering
\begin{tcolorbox}[
  width=\textwidth,
  colback=gray!5,
  colframe=gray,
  % equal height group=A, % Assigns this box to group 'A'
  title=Feedback with Running Errors
]
\small{
\begin{Verbatim}[breaklines,fontsize=\small]
prefix = "The above code is incorrect and got"
if metadata["error_code"] == -1:   # time limit exceeded
    feedback = f"{prefix} the following compilation error.\n"
    f"{metadata['error']}"
elif metadata["error_code"] == -2: # wrong answer
    feedback = f"{prefix} a wrong answer.\nInput: {metadata['inputs']}"
    f"\nGenerated Output: {metadata['output']}"
    f"\nExpected: {metadata['expected']}"
elif metadata["error_code"] == -3: # time limit exceeded
    feedback = f"{prefix} time limit exceeded.\n{metadata['error']}"
    f"\nInput: {metadata['inputs']}\nExpected: {metadata['expected']}"
    pass
elif metadata["error_code"] == -4: # runtime error
    if "error" in metadata.keys() and "inputs" in metadata.keys() \
    and "expected" in metadata.keys():
        feedback = f"{prefix} a runtime error.\n"
        f"Input: {metadata['inputs']}\n"
        f"Expected: {metadata['expected']}\n{metadata['error']}"
    else:
        feedback = f"{prefix} a runtime error.\n"
        f"{metadata['error_message']}"
elif metadata["error_code"] == -5:
    feedback = f"{prefix} the following test runner error.\n"
    f"{metadata['error']}"
feedback += "\nPlease re-answer the question "
            "based on the running errors.\n"
\end{Verbatim}
}
\end{tcolorbox}
\caption{Feedback prompt design when there are running errors in sequential sampling.}
\label{prompt_debug}
\end{figure}

\begin{figure}[htbp]
\centering
\includegraphics[width=\textwidth]{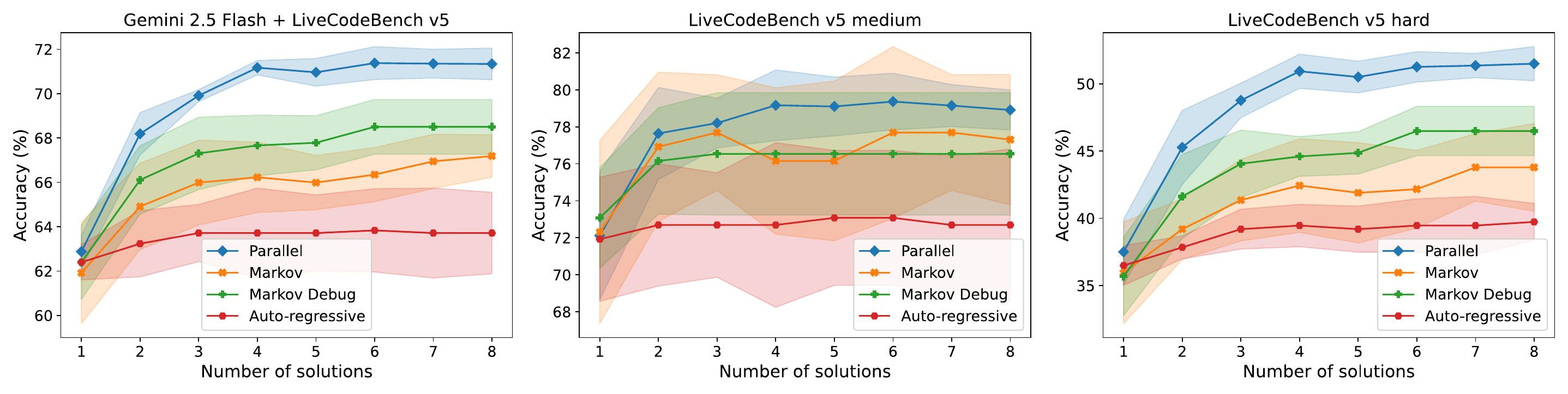}
\caption{Comparisons of LiveCodeBench v5 performance between parallel and sequential sampling using Gemini 2.5 Flash on (\emph{Left}) all questions; (\emph{Middle}) medium questions; (\emph{Right}) hard questions.\looseness=-1}
% \vspace{-0.2cm}
\label{code_compare2}
\end{figure}

\begin{figure}[htbp]
\centering
\includegraphics[width=\textwidth]{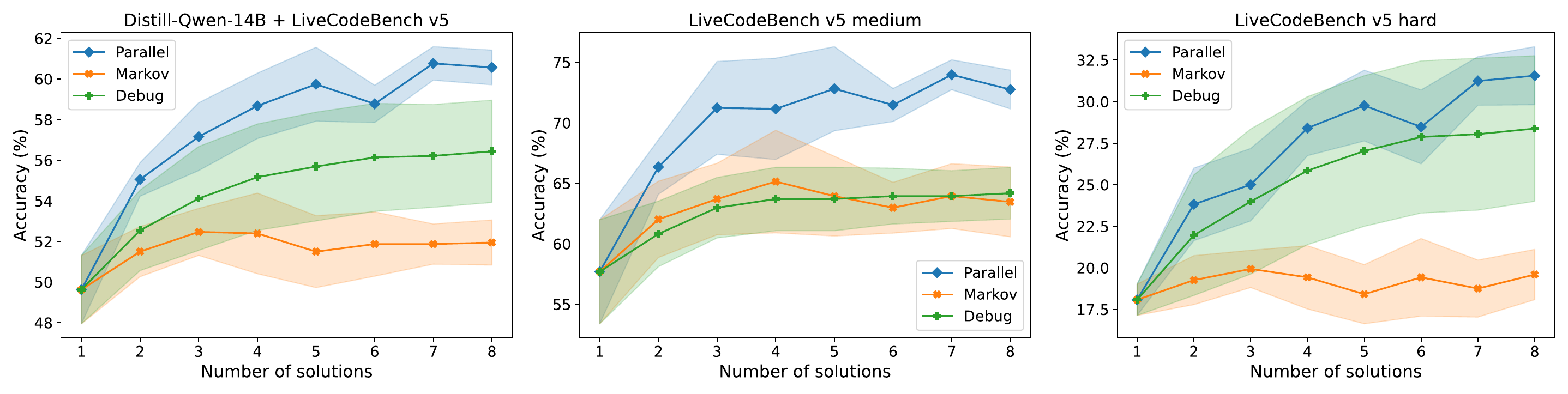}
\caption{Comparisons of LiveCodeBench v5 performance between parallel and sequential sampling using DeepSeek-R1-Distill Qwen-14B on (\emph{Left}) all questions; (\emph{Middle}) medium questions; (\emph{Right}) hard questions.\looseness=-1}
% \vspace{-0.2cm}
\label{code_compare3}
\end{figure}

\begin{figure}[htbp]
\centering
\includegraphics[width=\textwidth]{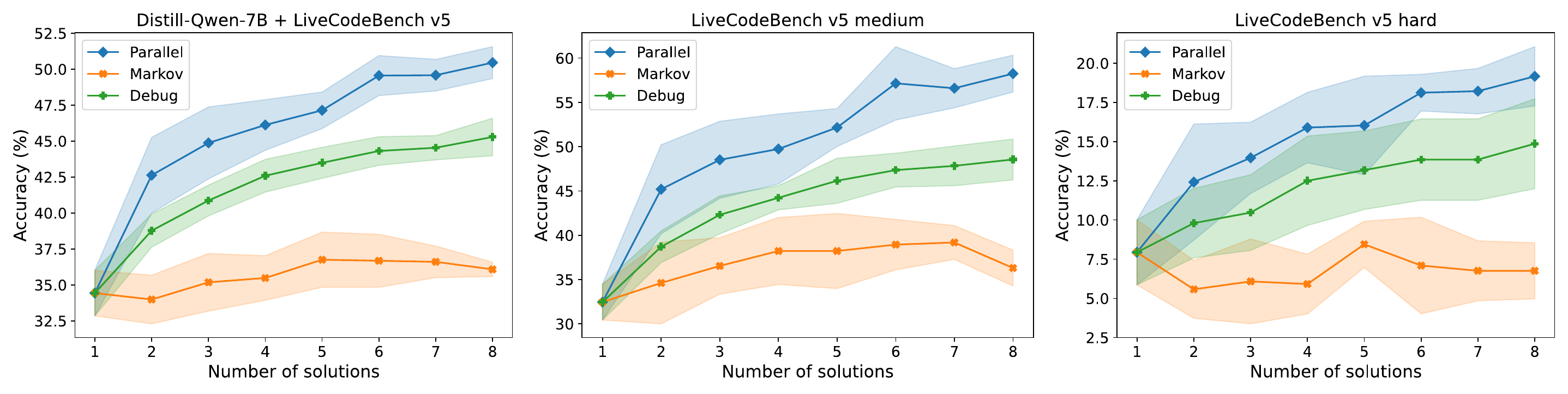}
\caption{Comparisons of LiveCodeBench v5 performance between parallel and sequential sampling using DeepSeek-R1-Distill Qwen-7B on (\emph{Left}) all questions; (\emph{Middle}) medium questions; (\emph{Right}) hard questions.\looseness=-1}
% \vspace{-0.2cm}
\label{code_compare4}
\end{figure}

\clearpage
\textbf{Empirical results in code generation task.} As shown in Figure~\ref{code_compare2}, \ref{code_compare3} and \ref{code_compare4}, we compare parallel and sequential sampling using various models on coding questions with different difficulty levels. Since these LRMs can achieve 95\% accuracy on easy questions, we only visualize the results from all questions, medium questions, and hard questions. We find that for Gemini 2.5 Flash, Markov sequential sampling with or without running errors can achieve comparable performance on medium questions, but not on hard questions. While for DeepSeek-R1-Distill Qwen models, running errors can bring significant performance gain on hard questions. We hypothesize that accuracy on hard questions is very low, and some hints from running errors are very useful.
\begin{figure}[htbp]
\centering
\begin{tcolorbox}[
  width=\textwidth,
  colback=gray!5,
  colframe=gray,
  % equal height group=A, % Assigns this box to group 'A'
  title=Generating test cases
]
\small{
Given a coding task, instead of providing the final script, your task is to generate a new test example (both input, output and explanation).
This is the problem:\newline\{Question\_content\}\newline
You need to provide a new test example. A good test example should be completely accurate and conform to the problem's format requirements, while also possessing enough discriminative power to distinguish correct code from incorrect code.
Before providing a test example, you must think carefully and reason step by step to derive an input and output you are very confident are correct. For example, start by designing an input you can reliably handle, then compute the output step by step. If you're unsure about the output, revise or re-design the input to ensure accuracy. Directly providing input/output pairs without this process is discouraged, as it often results in low accuracy.
Finally, after completing these previous thinking and derivation steps (you should not write the final test example unless you have gone through these steps very thoroughly), you MUST put your final test example in the following format:\newline
**Test Input:**\newline\verb|```|input here\verb|```|\newline\newline**Test Output:**\newline\verb|```|output here\verb|```|\newline\newline**Explanation:**\newline\newline
explanation here.\newline
\textcolor{blue}{Take care about the newline characters.}\newline
We already have \{Num\_Test\_Cases\} test samples:\newline
Example 1:\newline**Test Input:**\newline\verb|```|\{Input1\}\verb|```|\newline\newline**Test Output:**\newline\verb|```|\{Output2\}\verb|```|\newline\newline
...
}
\end{tcolorbox}
\caption{Prompt design for generating test cases based on public tests.}
\label{prompt_gen_tests}
\end{figure}

\section{Empirical Studies on Understanding the Performance Gap}\label{appendix_ablation}

\subsection{Aggregation}

\begin{figure}[htbp]
\centering
\includegraphics[width=\textwidth]{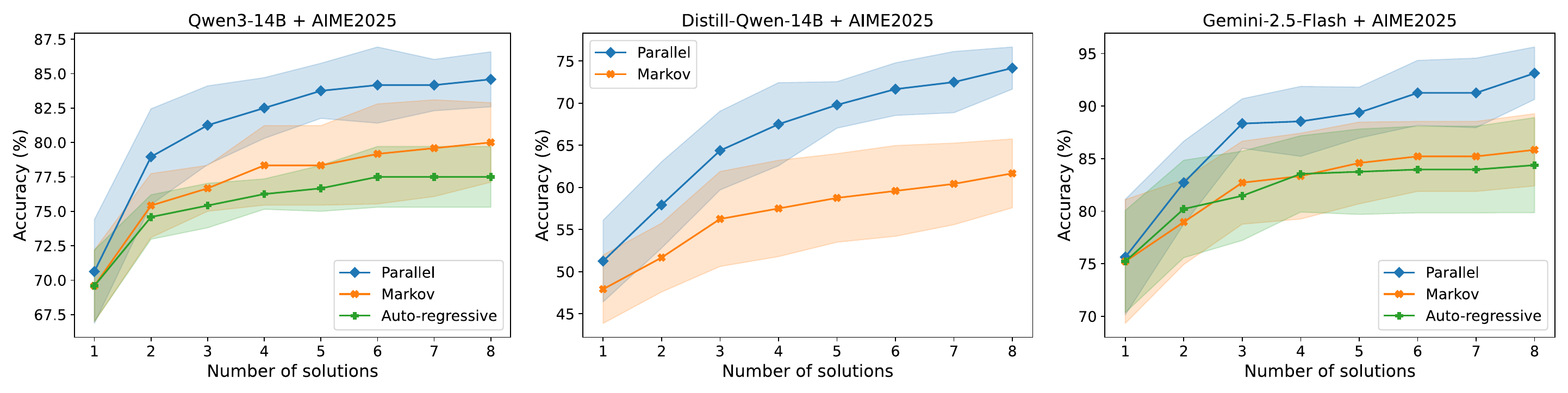}
\caption{Comparisons of AIME2025 performance between parallel and sequential sampling using (\emph{Left}) Qwen3-14B; (\emph{Middle}) DeepSeek-R1-Distill Qwen-14B; (\emph{Right}) Gemini 2.5 Flash. Both sampling approaches use the ideal best-of-N aggregation with the assumption of perfect verifier.\looseness=-1}
% \vspace{-0.2cm}
\label{math_aggregation}
\end{figure}

For math competition task, besides the majority voting aggregation adopted in the main paper, we also attempt best-of-N aggregation with a perfect verifier. As shown in Figure~\ref{math_aggregation}, we find that under such an aggregation operator, the performance gap between parallel and sequential sampling still exists, even larger than the case of majority voting.

\begin{figure}[htbp]
\centering
\includegraphics[width=\textwidth]{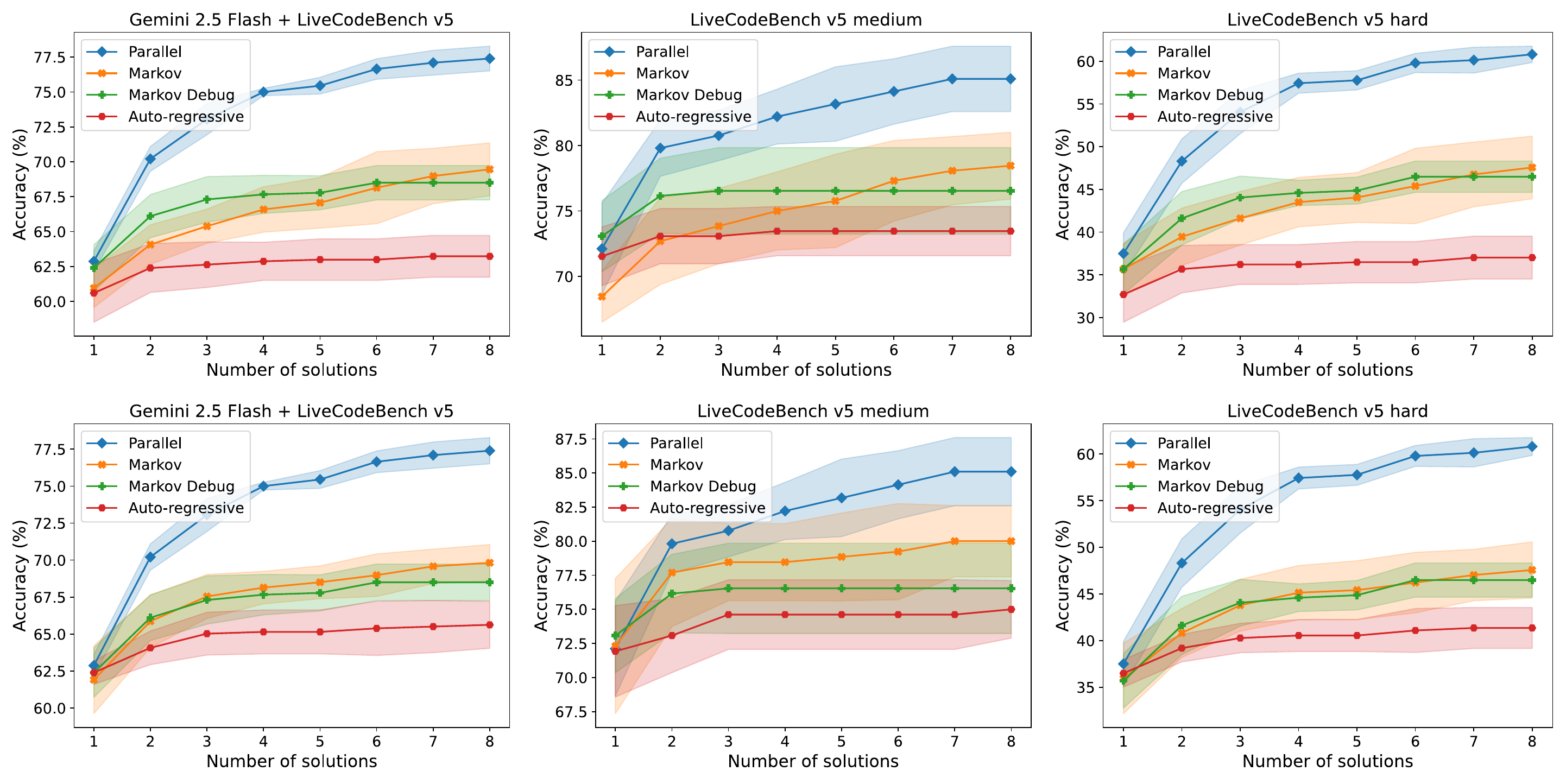}
\caption{Comparisons (with different difficulty levels) of LiveCodeBench v5 performance between parallel and sequential sampling using Gemini 2.5 Flash. Both sampling approaches adopt the same best-of-N aggregation based on rewards on public and private tests. In sequential sampling, we apply feedback prompt of (\emph{Top}) ``Please re-answer'' and (\emph{Bottom}) self-refinementing.\looseness=-1}
% \vspace{-0.2cm}
\label{code_aggregation2_gemini-2.5-flash}
\end{figure}

\begin{figure}[htbp]
\centering
\includegraphics[width=\textwidth]{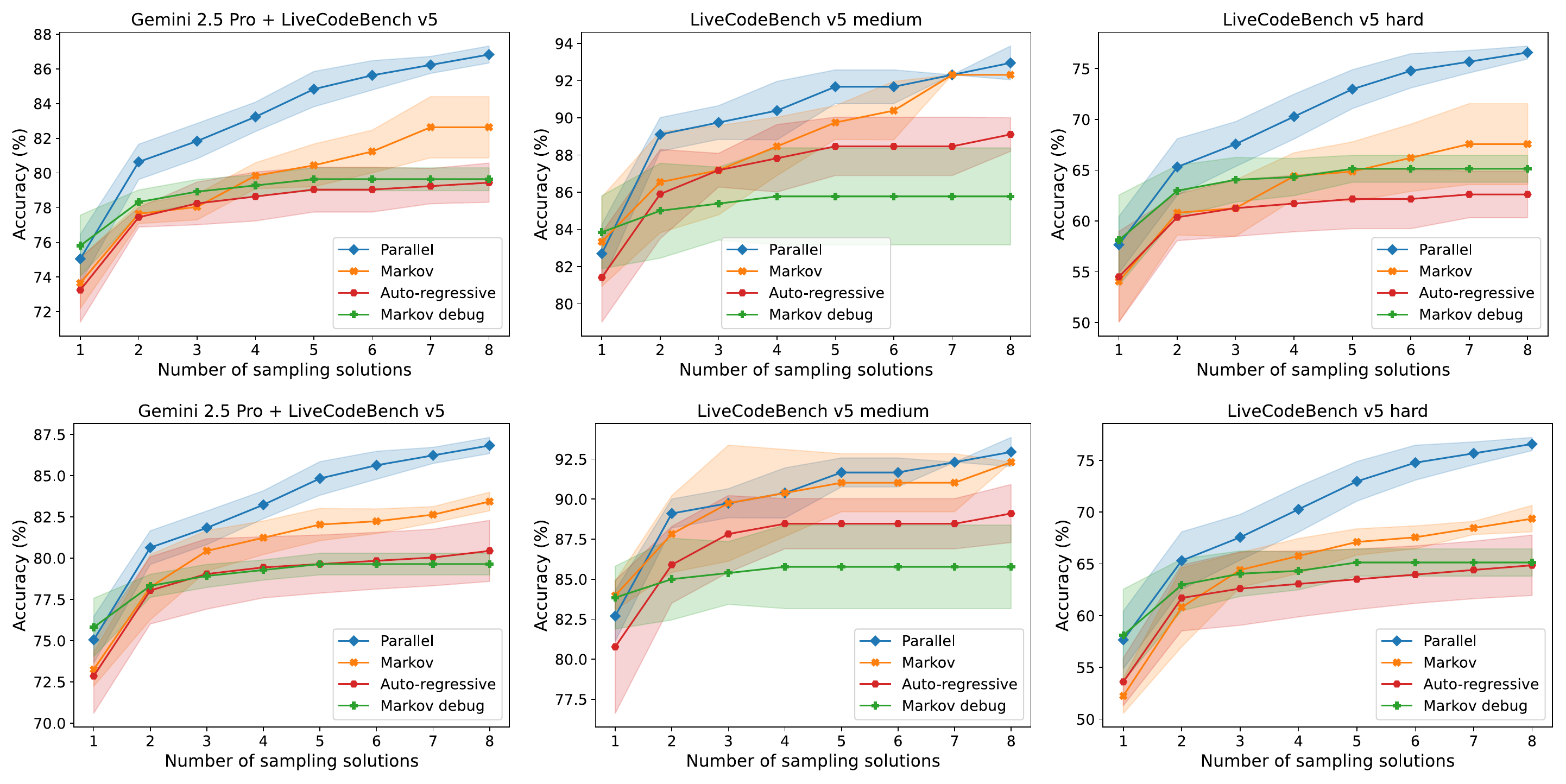}
\caption{Comparisons (with different difficulty levels) of LiveCodeBench v5 performance between parallel and sequential sampling using Gemini 2.5 Pro. Both sampling approaches adopt the same best-of-N aggregation based on rewards on public and private tests. In sequential sampling, we apply feedback prompt of (\emph{Top}) ``Please re-answer'' and (\emph{Bottom}) self-refinementing.\looseness=-1}
% \vspace{-0.2cm}
\label{code_aggregation2_gemini-2.5-pro}
\end{figure}

\begin{figure}[htbp]
\centering
\includegraphics[width=\textwidth]{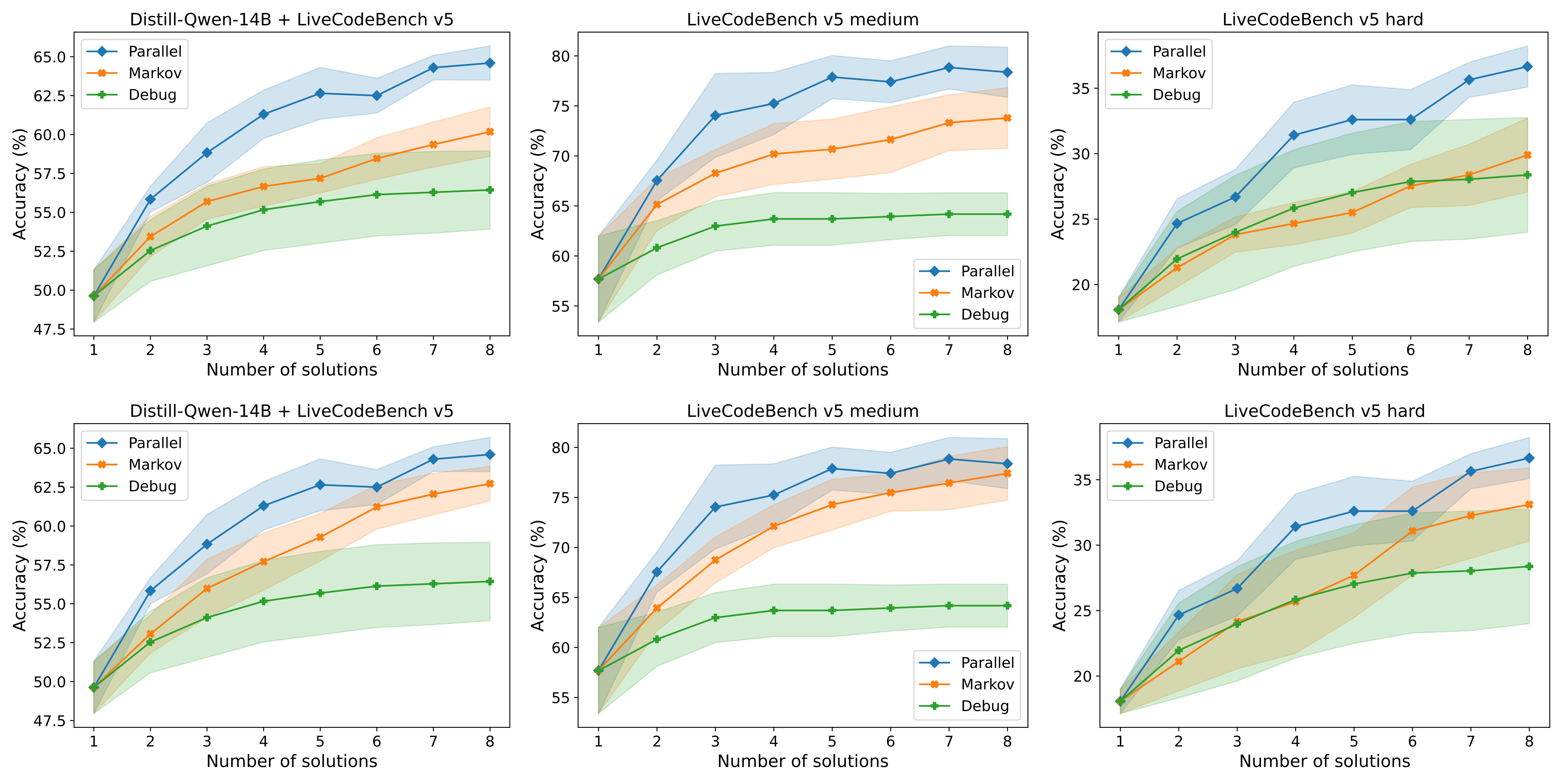}
\caption{Comparisons (with different difficulty levels) of LiveCodeBench v5 performance between parallel and sequential sampling using DeepSeek-R1-Distill Qwen-14B. Both sampling approaches adopt the same best-of-N aggregation based on rewards on public and private tests. In sequential sampling, we apply feedback prompt of (\emph{Top}) ``Please re-answer'' and (\emph{Bottom}) self-refinementing.\looseness=-1}
% \vspace{-0.2cm}
\label{code_aggregation2_distill-14b}
\end{figure}

\begin{figure}[htbp]
\centering
\includegraphics[width=\textwidth]{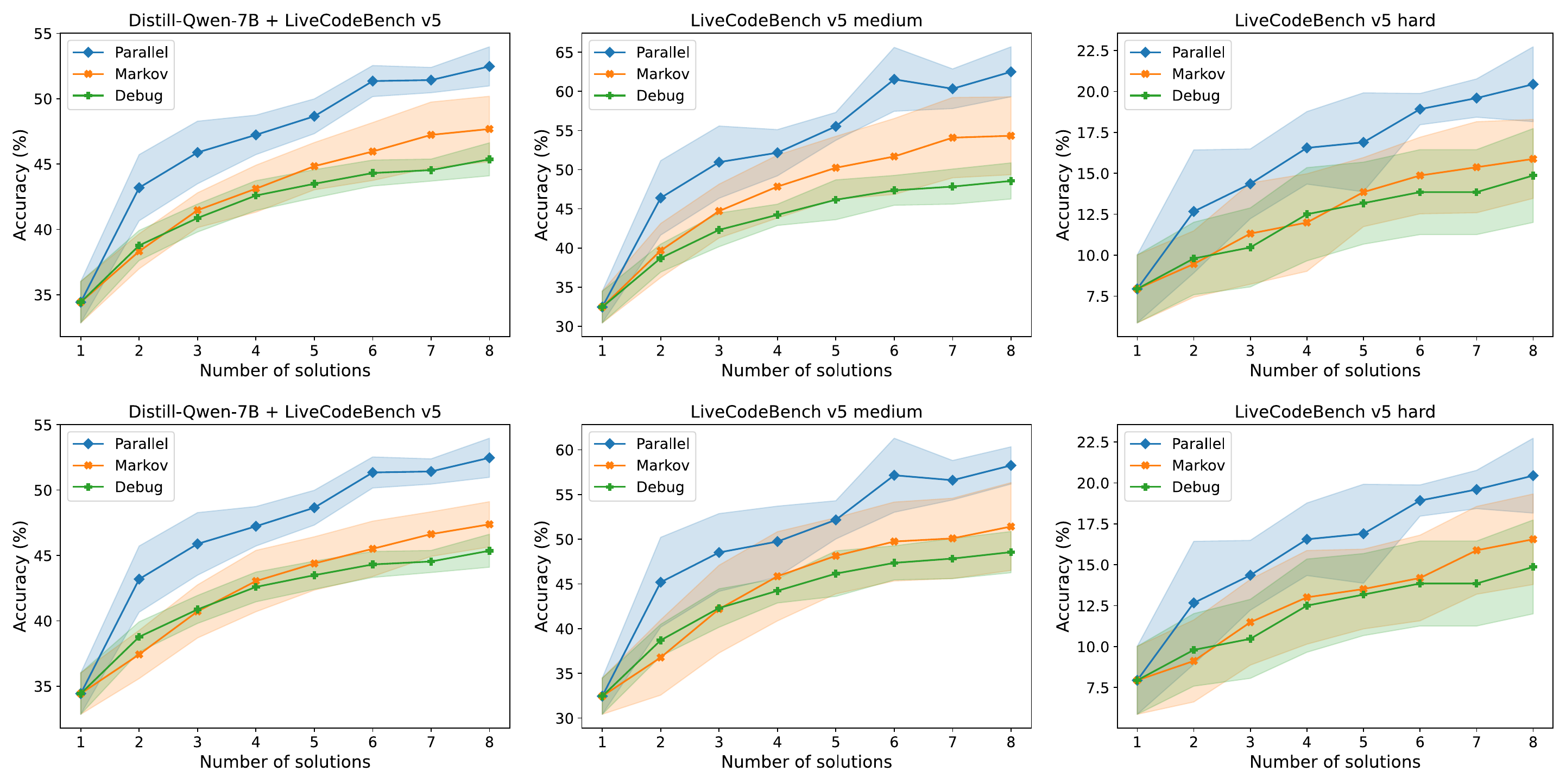}
\caption{Comparisons (with different difficulty levels) of LiveCodeBench v5 performance between parallel and sequential sampling using DeepSeek-R1-Distill Qwen-7B. Both sampling approaches adopt the same best-of-N aggregation based on rewards on public and private tests. In sequential sampling, we apply feedback prompt of (\emph{Top}) ``Please re-answer'' and (\emph{Bottom}) self-refinementing.\looseness=-1}
% \vspace{-0.2cm}
\label{code_aggregation2_distill-7b}
\end{figure}

For code generation task, we show more results on questions with different difficulty levels, such Figure~\ref{code_aggregation2_gemini-2.5-flash} (Gemini 2.5 Flash), Figure~\ref{code_aggregation2_gemini-2.5-pro} (Gemini 2.5 Pro), Figure~\ref{code_aggregation2_distill-14b} (DeepSeek-R1-Distill Qwen-14B), and Figure~\ref{code_aggregation2_distill-7b} (DeepSeek-R1-Distill Qwen-7B). Both parallel and sequential sampling adopt the same best-of-N aggregation operator with rewarding on public and private tests. For most LRMs, different feedback prompts do not result in much performance difference. Only for DeepSeek-R1-Distill Qwen-14B, after aggregation, self-refinement feedback works better. However, overall, aggregation on sequential sampling can narrow but not close the performance gap.

\subsection{Extended input context}

\begin{figure}[htbp]
\centering
\includegraphics[width=\textwidth]{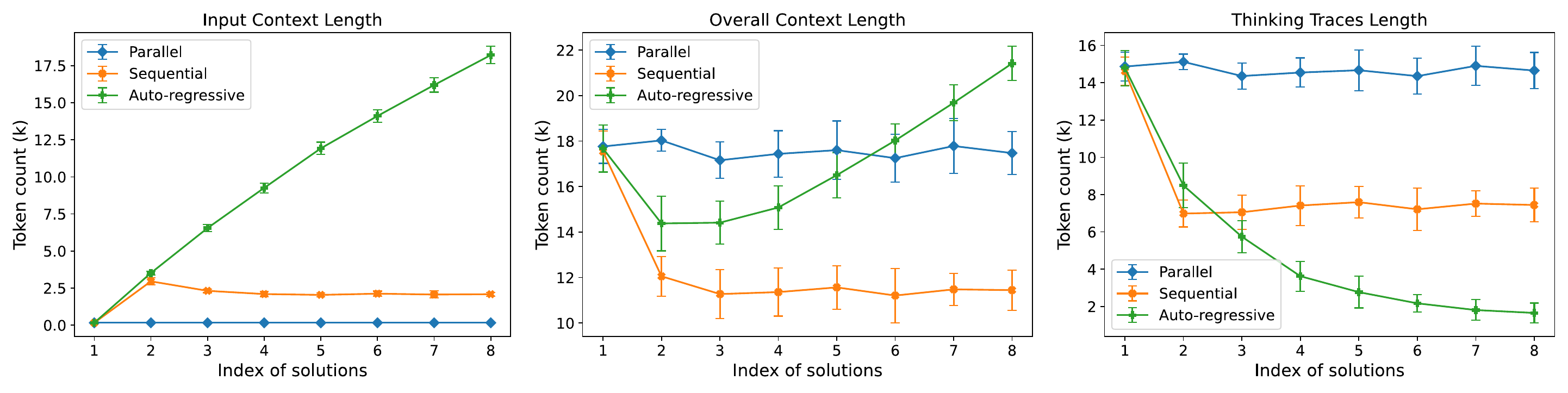}
\caption{(\emph{Left}) input context length, (\emph{Middle}) overall context length, (\emph{Right}) thinking traces length when using Gemini 2.5 Flash on AIME2025 under parallel and sequential sampling.\looseness=-1}
% \vspace{-0.2cm}
\label{context_len_math_gemini-2.5-flash}
\end{figure}

\begin{figure}[htbp]
\centering
\includegraphics[width=\textwidth]{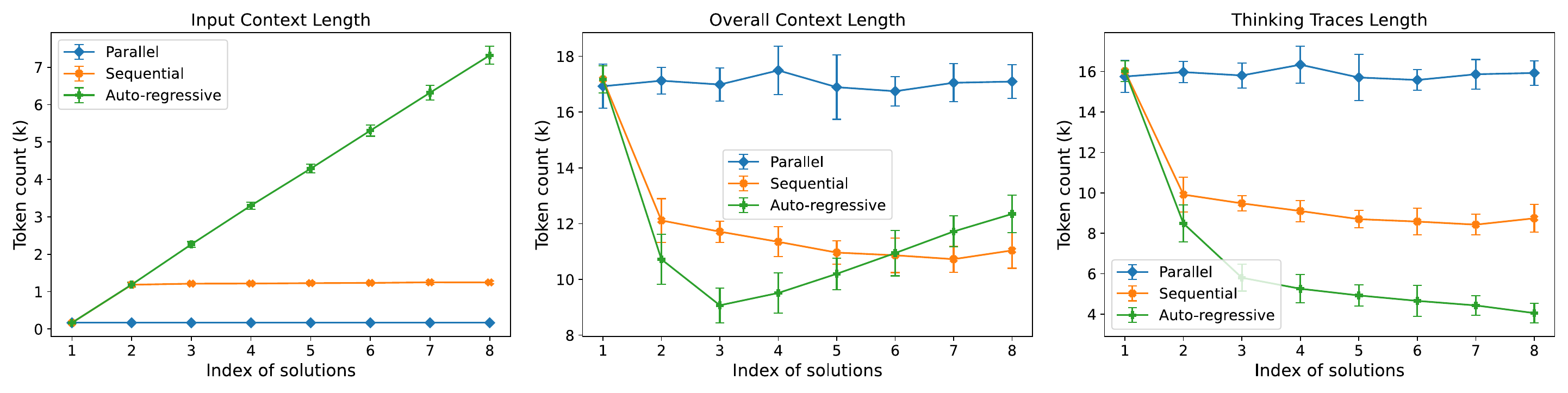}
\caption{(\emph{Left}) input context length, (\emph{Middle}) overall context length, (\emph{Right}) thinking traces length when using Qwen3-14B on AIME2025 under parallel and sequential sampling.\looseness=-1}
% \vspace{-0.2cm}
\label{context_len_math_qwen3-14b}
\end{figure}

\begin{figure}[htbp]
\centering
\includegraphics[width=\textwidth]{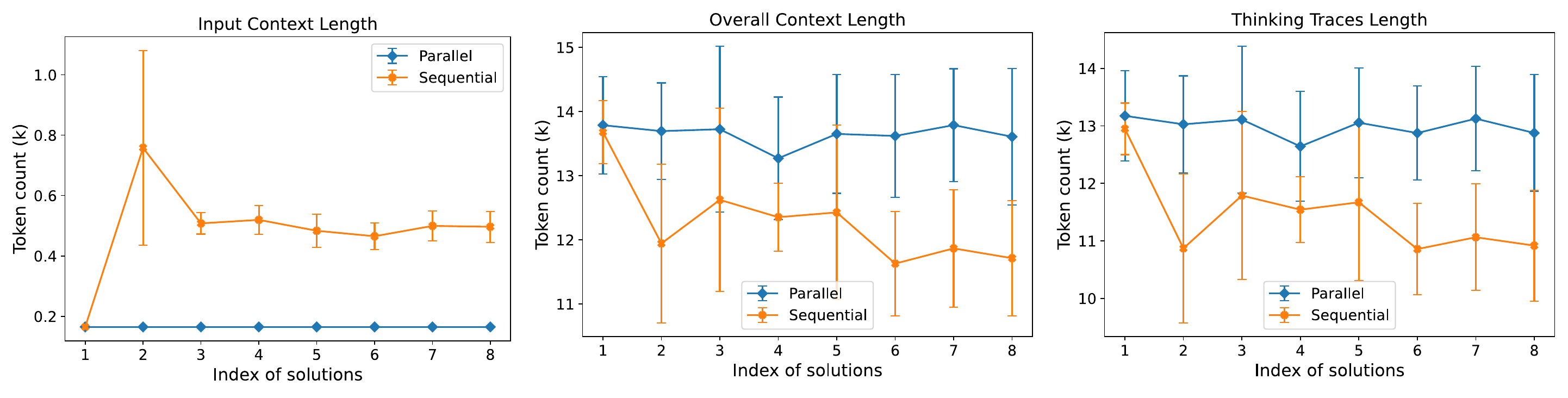}
\caption{(\emph{Left}) input context length, (\emph{Middle}) overall context length, (\emph{Right}) thinking traces length when using DeepSeek-R1-Distill Qwen-14B on AIME2025 under parallel and sequential sampling.\looseness=-1}
% \vspace{-0.2cm}
\label{context_len_math_distill-14b}
\end{figure}

\begin{figure}[htbp]
\centering
\includegraphics[width=\textwidth]{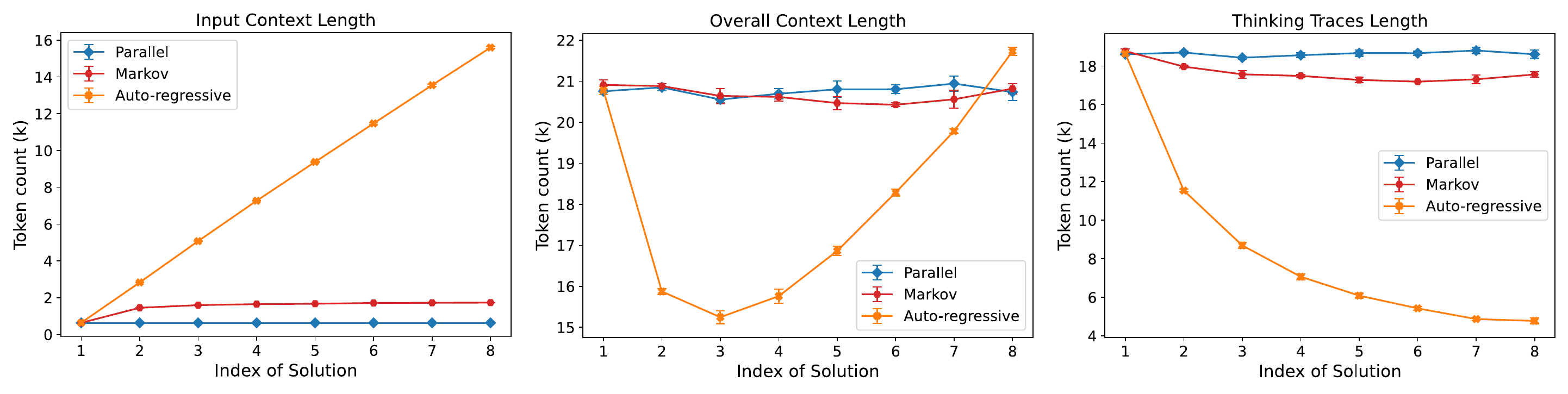}
\caption{(\emph{Left}) input context length, (\emph{Middle}) overall context length, (\emph{Right}) thinking traces length when using Gemini 2.5 Pro on LiveCodeBench v5 under parallel/sequential sampling.\looseness=-1}
% \vspace{-0.2cm}
\label{context_len_code_gemini-2.5-pro}
\end{figure}

We visualize the results of AIME2025, including Gemini 2.5 Flash in Figure~\ref{context_len_math_gemini-2.5-flash}, Qwen3-14B in Figure~\ref{context_len_math_qwen3-14b}, DeepSeek-R1-Distill Qwen-14B in Figure~\ref{context_len_math_distill-14b}, and Gemini 2.5 Pro on LiveCodeBench in Figure~\ref{context_len_code_gemini-2.5-pro}. These results consistently show there is no significant context differences between parallel and sequential sampling. Additionally, LRMs demonstrate ``laziness'' in sequential sampling.\looseness=-1

\begin{figure}[htbp]
\centering
\includegraphics[width=\textwidth]{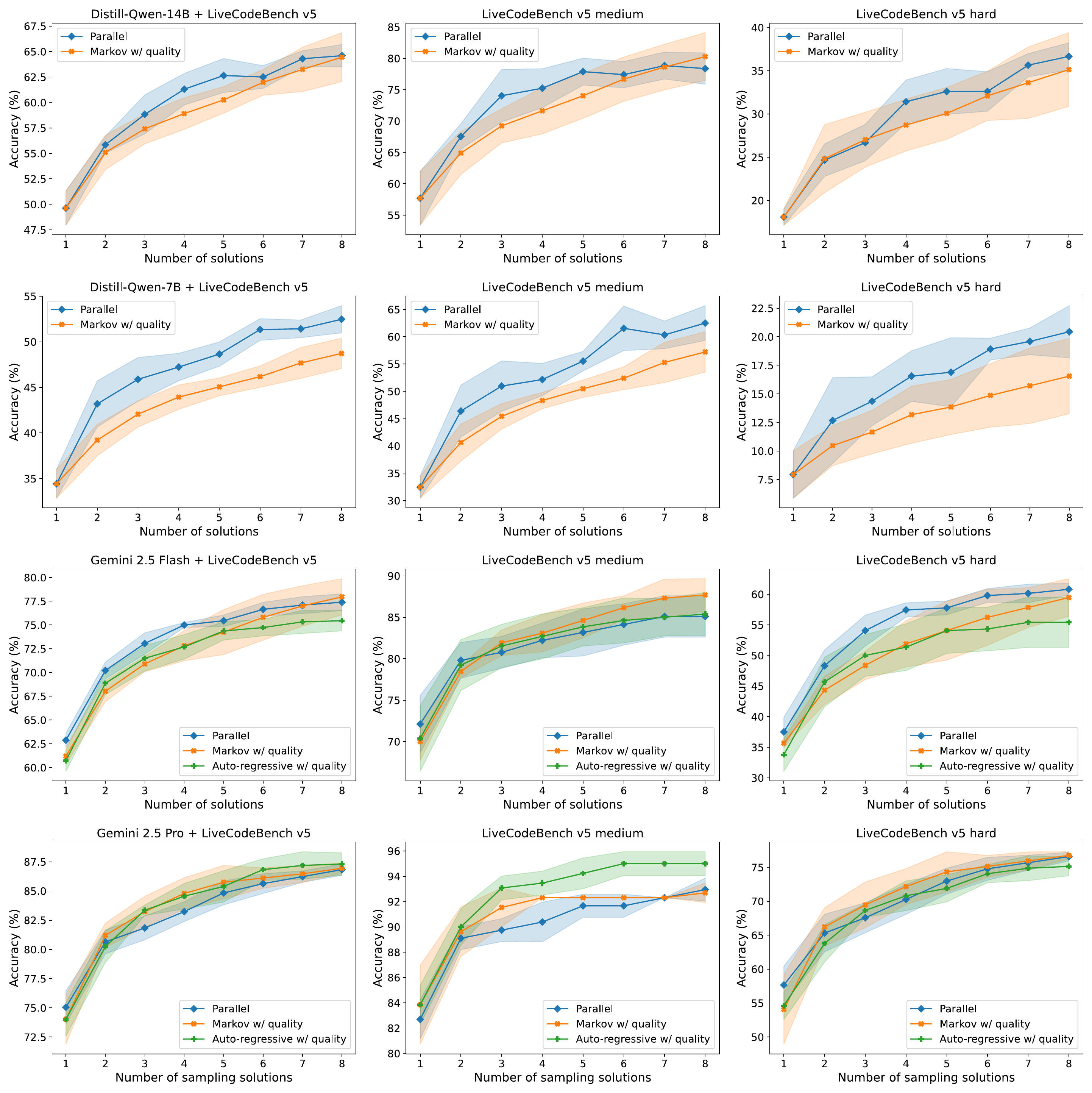}
\caption{Comparisons (with different difficulty levels) of LiveCodeBench v5 performance of (from \emph{Top} to \emph{Bottom}) DeepSeek-R1-Distill Qwen-14B/7B, Gemini 2.5 Flash/Pro under parallel sampling and sequential sampling using the best-of-N aggregation. Each solution is rewarded based on public and private tests. For sequential sampling, the running errors on public and private tests are provided as high-quality guidance.\looseness=-1}
% \vspace{-0.2cm}
\label{high-quality-levels}
\end{figure}

% \clearpage

\subsection{Solution exploration}

\textbf{``Laziness'' of LRMs in sequential sampling.} When solving the question~\footnote{\url{https://atcoder.jp/contests/arc184/tasks/arc184_e}} from LiveCodeBench using Gemini 2.5 Pro, the 5-th, 7-th, and 8-th solutions contain the same generated code program, as shown in Figure~\ref{laziness1}, \ref{laziness2}, and \ref{laziness3}. The LRM gets stuck by the same public test case and keeps receiving the wrong answer error as feedback in the sequential chain, and then being lazy to explore alternatives.\looseness=-1

\textbf{High-quality feedback for failure.} As shown in  Figure~\ref{high-quality-levels}, running errors on both public and private tests (high-quality feedback) significantly improve performance of sequential sampling, especially for medium questions. For hard questions, parallel sampling still shows superiority, mainly for Distill-Qwen-14B/7B and Gemini 2.5 Flash.  

% \textbf{Measuring solution similarity.}

\begin{figure}[t]
\centering
\begin{tcolorbox}[
  width=\textwidth,
  colback=gray!5,
  colframe=gray,
  equal height group=Z, % Assigns this box to group 'A'
]
\scriptsize{
\begin{Verbatim}[breaklines,fontsize=\small]
# YOUR CODE HERE
import sys

sys.setrecursionlimit(2 * 10**6)

MOD = 998244353
memo_f = {}
memo_g = {}
memo_t_op = {}
memo_u_op = {}

def T_op_group(group, m):
    state = (group, m)
    if state in memo_t_op:
        return memo_t_op[state]
    
    res = []
    for seq in group:
        new_seq = []
        current_sum = 0
        for bit in seq:
            current_sum = (current_sum + int(bit)) % 2
            new_seq.append(str(current_sum))
        res.append("".join(new_seq))
    
    result = tuple(sorted(res))
    memo_t_op[state] = result
    return result

def U_op_group(group, m):
    state = (group, m)
    if state in memo_u_op:
        return memo_u_op[state]
    
    res = []
    for seq in group:
        new_seq = []
        current_sum = 0
        for bit in seq:
            current_sum = (current_sum + int(bit)) % 2
            new_seq.append(str(1 - current_sum))
        res.append("".join(new_seq))

    result = tuple(sorted(res))
    memo_u_op[state] = result
    return result
\end{Verbatim}
}
\end{tcolorbox}
\caption{Solution generated by Gemini 2.5 Pro repeatedly appear in the auto-regressive sequential sampling when solving ``Accumulating Many Times'' from LiveCodeBench.\looseness=-1}
\label{laziness1}
\end{figure}

\begin{figure}[t]
\centering
\begin{tcolorbox}[
  width=\textwidth,
  colback=gray!5,
  colframe=gray,
  equal height group=A, % Assigns this box to group 'A'
]
\scriptsize{
\begin{Verbatim}[breaklines,fontsize=\small]
def solve_f(group, m):
    if len(group) <= 1 or m == 0:
        return 0
    
    state = (group, m)
    if state in memo_f:
        return memo_f[state]

    g0 = [s for s in group if s[0] == '0']
    g1 = [s for s in group if s[0] == '1']
    
    g0_suffs = tuple(sorted([s[1:] for s in g0])) if m > 1 else tuple()
    g1_suffs = tuple(sorted([s[1:] for s in g1])) if m > 1 else tuple()
    
    res = solve_f(g0_suffs, m - 1)
    
    len_g1 = len(g1)
    if len_g1 > 1:
        term_g1 = (len_g1 * (len_g1 - 1) // 2) % MOD
        res = (res + term_g1) % MOD
    
    if g1_suffs:
        res = (res + solve_g(g1_suffs, m - 1)) % MOD
    
    memo_f[state] = res
    return res
\end{Verbatim}
}
\end{tcolorbox}
\caption{(Continued) solution generated by Gemini 2.5 Pro repeatedly appear in the auto-regressive sequential sampling when solving ``Accumulating Many Times'' from LiveCodeBench.\looseness=-1}
\label{laziness2}
\end{figure}

\begin{figure}[t]
\centering
\begin{tcolorbox}[
  width=\textwidth,
  colback=gray!5,
  colframe=gray,
  % equal height group=A, % Assigns this box to group 'A'
]
\scriptsize{
\begin{Verbatim}[breaklines,fontsize=\small]
def solve_g(group, m):
    if len(group) <= 1 or m == 0:
        return 0
    
    state = (group, m)
    if state in memo_g:
        return memo_g[state]

    g0 = [s for s in group if s[0] == '0']
    g1 = [s for s in group if s[0] == '1']

    res = 0
    len_g0 = len(g0)
    if len_g0 > 1:
        res = (res + len_g0 * (len_g0 - 1) // 2) % MOD
    
    len_g1 = len(g1)
    if len_g1 > 1:
        res = (res + len_g1 * (len_g1 - 1) // 2) % MOD
    
    if m > 1:
        g0_suffs = tuple(sorted([s[1:] for s in g0]))
        g1_suffs = tuple(sorted([s[1:] for s in g1]))

        if g1_suffs:
            f_world_suffs = T_op_group(g1_suffs, m - 1)
            res = (res + solve_f(f_world_suffs, m - 1)) % MOD
        
        if g0_suffs:
            g_world_suffs = U_op_group(g0_suffs, m - 1)
            res = (res + solve_g(g_world_suffs, m - 1)) % MOD

    memo_g[state] = res
    return res

def main():
    try:
        readline = sys.stdin.readline
        N, M = map(int, readline().split())
        sequences = tuple(sorted(readline().strip().replace(' ', '') for _ in range(N)))
    except (IOError, ValueError):
        return

    print(solve_f(sequences, M))

if __name__ == "__main__":
    main()
\end{Verbatim}
}
\end{tcolorbox}
\caption{(Continued) solution generated by Gemini 2.5 Pro repeatedly appear in the auto-regressive sequential sampling when solving ``Accumulating Many Times'' from LiveCodeBench.\looseness=-1}
\label{laziness3}
\end{figure}